\documentclass[10pt,twocolumn,letterpaper]{article}

\usepackage[pagenumbers]{cvpr}

\definecolor{cvprblue}{rgb}{0.21,0.49,0.74}
\usepackage[pagebackref,breaklinks,colorlinks,allcolors=cvprblue]{hyperref}

\usepackage{booktabs}
\usepackage{multirow}
\usepackage[table]{xcolor}
\usepackage{colortbl}
\usepackage{algorithm}
\usepackage{algpseudocode}

\def\paperID{17372} 
\def\confName{CVPR}
\def\confYear{2026}

\title{SliderEdit: Continuous Image Editing with Fine-Grained Instruction Control}

\author{Arman Zarei\textsuperscript{1},\hspace{2pt} Samyadeep Basu\textsuperscript{2},\hspace{2pt} Mobina Pournemat\textsuperscript{1},\hspace{2pt} Sayan Nag\textsuperscript{2},\hspace{2pt} Ryan Rossi\textsuperscript{2},\hspace{2pt} Soheil Feizi\textsuperscript{1}\\
  \textsuperscript{1}University of Maryland \hspace{14pt} \textsuperscript{2}Adobe Research
}

\begin{document}
\twocolumn[{%
    \renewcommand\twocolumn[2][]{#1}
    \maketitle
    \iftoggle{cvprfinal}{\vspace{-0.7cm}}{\vspace{-1cm}}
    \begin{center}
        \centering \centering
        \includegraphics[width=0.96\textwidth]{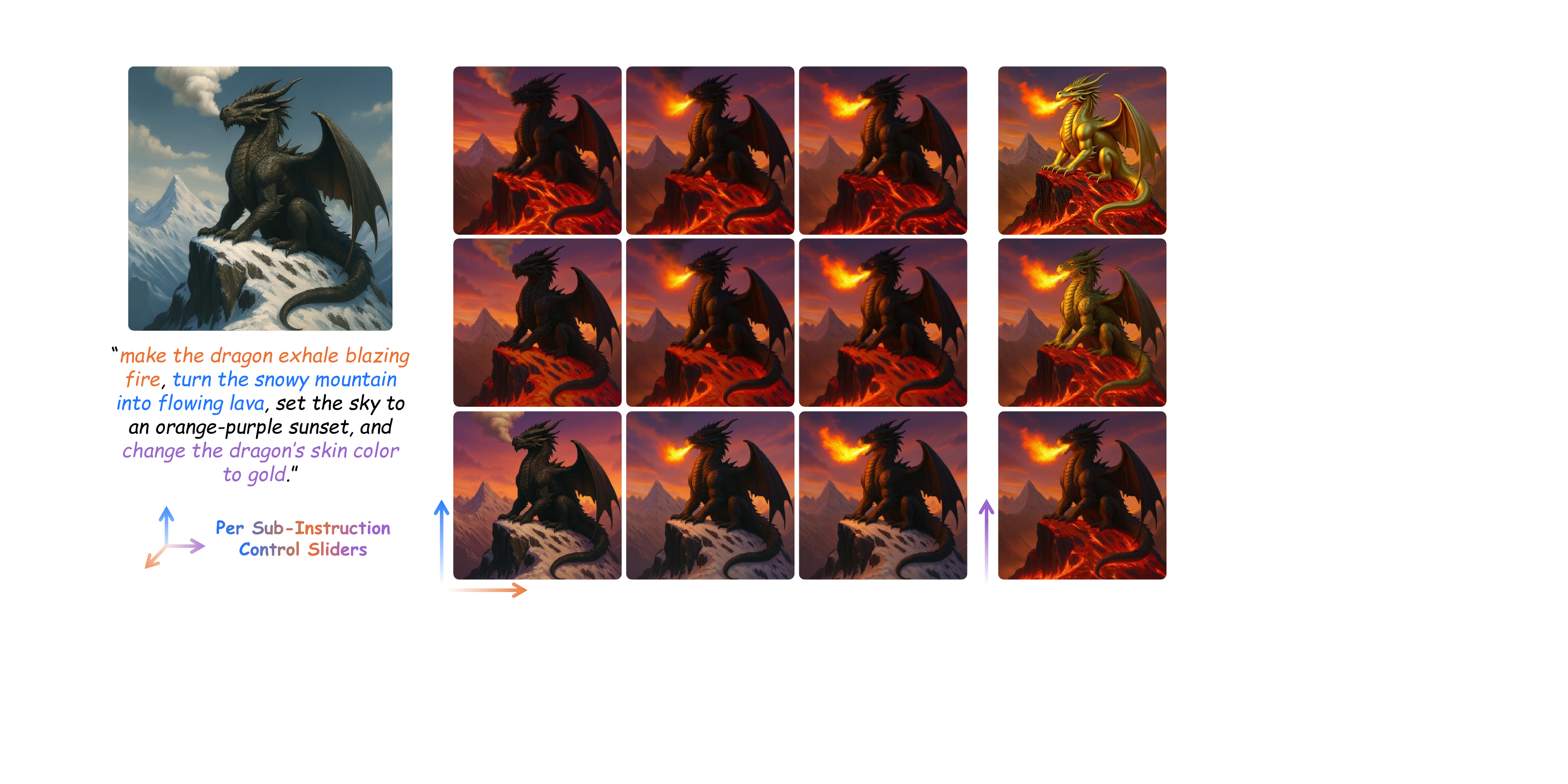}
        \vspace{-0.2cm}
        \captionof{figure}{\textbf{SliderEdit produces continuous edit trajectories in state-of-the-art instruction-based image editing models.} Our method provides fine-grained and disentangled control over the intensity of edit attributes described in an instruction, allowing continuous transitions between editing strengths. Despite its effectiveness, SliderEdit is extremely lightweight and can be trained efficiently to transform a state-of-the-art instruction-based image editing model into a continuously controllable editing framework.}
    \label{fig:teaser}
    \end{center}
}] 

\begin{abstract}
Instruction-based image editing models have recently achieved impressive performance, enabling complex edits to an input image from a multi-instruction prompt. However, these models apply each instruction in the prompt with a fixed strength, limiting the user’s ability to precisely and continuously control the intensity of individual edits.
We introduce \textbf{SliderEdit}, a framework for continuous image editing with fine-grained, interpretable instruction control. Given a multi-part edit instruction, SliderEdit disentangles the individual instructions and exposes each as a globally trained slider, allowing smooth adjustment of its strength. Unlike prior works that introduced slider-based attribute controls in text-to-image generation, typically requiring separate training or fine-tuning for each attribute or concept, our method learns a \emph{single} set of low-rank adaptation matrices that generalize across diverse edits, attributes, and compositional instructions. This enables continuous interpolation along individual edit dimensions while preserving both spatial locality and global semantic consistency. We apply SliderEdit to state-of-the-art image editing models, including FLUX-Kontext and Qwen-Image-Edit, and observe substantial improvements in edit controllability, visual consistency, and user steerability. 
To the best of our knowledge, we are the first to explore and propose a framework for continuous, fine-grained instruction control in instruction-based image editing models. Our results pave the way for interactive, instruction-driven image manipulation with continuous and compositional control.\iftoggle{cvprfinal}{\footnote[1]{Project page is available at: \href{https://armanzarei.github.io/SliderEdit}{https://armanzarei.github.io/SliderEdit}}}

\end{abstract}
    
\vspace{-0.5cm}
\section{Introduction}
\label{sec:intro}


Recent advances in large-scale diffusion~\cite{ho2020denoising, nichol2021improved, rombach2022high} and flow-matching models~\cite{lipman2022flow, esser2024scaling} have revolutionized image synthesis, enabling unprecedented photorealism and semantic fidelity. Building on these foundations, instruction-based image editing has emerged as a powerful paradigm, allowing users to modify images through natural language commands~\cite{instructpix2pix, icedit, qwenimageedit, fluxkontext}. The latest state-of-the-art models, such as \emph{FLUX-Kontext}~\cite{fluxkontext} and \emph{Qwen-Image-Edit}~\cite{qwenimageedit}, can perform a wide spectrum of manipulations, from global scene and style transformations to highly localized, fine-grained edits, all within a unified text-driven framework.

Despite these advances, current instruction-based editing models remain inherently \emph{discrete}: they apply edits in an all-or-nothing manner, offering limited control over how strongly each instruction is expressed. For example, given an image of a dragon and a multi-instruction prompt such as “change the skin color to gold and make it exhale fire”, existing models generate a single fixed outcome for a given prompt. While doing multiple generations may yield different variations, it does not allow systematic adjustment of individual edit \emph{strengths}, such as turning the skin slightly gold versus bright metallic gold, or adding a small flame versus a large burst of fire. This lack of fine-grained, continuous control limits both user flexibility and interpretability—the two key properties for truly interactive image editing.


To address this gap, we propose \emph{SliderEdit}, a framework for continuous image editing with fine-grained instruction control. Our goal is to extend state-of-the-art instruction-based editing models into systems that support \emph{continuous, disentangled, and interpretable control} over the effects of individual editing instructions. Specifically, given a multi-instruction prompt, SliderEdit assigns each instruction its own slider, allowing smooth adjustment of its influence between suppression, full application, and amplification (See Fig. \ref{fig:teaser}). These sliders provide intuitive and flexible control over complex multi-instruction edits, operating seamlessly without any re-training or per-instruction fine-tuning.

Our key insight is that the latent representations of modern multimodal diffusion transformers (MMDiTs) encode instruction semantics within localized token embeddings. By identifying and selectively modulating these tokens, we can gain fine-grained control over how individual instructions affect the output. Building on this observation, \emph{SliderEdit} employs a small set of learnable low-rank adaptation matrices that act directly on instruction-relevant token embeddings. These adapters are trained using a novel and lightweight objective, the \emph{Partial Prompt Suppression (PPS)} loss, which teaches the model how to suppress or neutralize the visual effect of a specific instruction. The loss simply requires that the model's output, when given the full prompt, matches the output produced when the target instruction is removed, making it intuitive, interpretable, and easy to optimize. Once trained, these low-rank adapters naturally yield continuous sliders by smoothly scaling their learned weights, enabling interpretable adjustment of each instruction's influence. For single-instruction edits, we further extend this idea by applying the adapter across all image and text tokens, resulting in smoother edit trajectories. 

\emph{SliderEdit} integrates seamlessly with existing state-of-the-art instruction-based image editing models such as \emph{FLUX-Kontext} and \emph{Qwen-Image-Edit}, requiring only minimal additional training. Our approach provides a unified framework for continuous and compositional control across diverse editing scenarios—from subtle attribute adjustments and stylistic refinements to complex, multi-object scene manipulations. Through both quantitative evaluation and qualitative analyses, we show that SliderEdit delivers superior edit controllability and semantic disentanglement.

In summary, our main contributions are:
\begin{itemize}
\item We are the \emph{first to explore and propose a framework for continuous instruction-based image editing}, enabling smooth, fine-grained, and interpretable modulation of edit intensity for individual instructions.
\item We propose \emph{Partial Prompt Suppression} loss, which enables efficient training of instruction-aware adapters that learn disentangled, continuous control over edit strengths.
\item We demonstrate seamless integration of our framework with state-of-the-art foundation image editing models, achieving substantial improvements in edit consistency and user controllability.
\end{itemize}
\section{Related Works}

\subsection{Image Editing}

Image editing methods have advanced rapidly in recent years. Early approaches built on diffusion priors enabled flexible editing by perturbing and denoising input images~\cite{sdedit}. Subsequent methods~\cite{pnpinversion, prompt2prompt, imagic, stableflow, rfedit} formulated editing as steering the diffusion trajectory through optimization or conditioning while preserving image fidelity. With the emergence of instruction-based editing~\cite{instructpix2pix, icedit, promptartisan, foi, zone}, models began to directly interpret natural language commands, allowing intuitive user control. More recently, large foundation models for instruction-based image editing~\cite{qwenimageedit, fluxkontext} have achieved remarkable versatility, performing both local and global modifications within a unified architecture. Despite their impressive capabilities, these models lack \emph{fine-grained controllability}, i.e., the ability to continuously adjust the strength of individual edits. Our work addresses this limitation through a framework that enables continuous and interpretable instruction-level control.


\subsection{Continuous Attribute Slider}

A growing body of work in image generative modeling has explored continuous attribute control over generated images. Before diffusion models, much of this effort focused on learning structured and manipulable latent spaces in GANs and VAEs~\cite{härkönen2020ganspacediscoveringinterpretablegan, karras2019stylebasedgeneratorarchitecturegenerative, shen2020interpretinglatentspacegans, Abdal_2021, hou2024deepfeatureconsistentvariational}. These approaches discovered semantically meaningful directions in the latent space that correspond to interpretable visual attributes. 
With the emergence of text-to-image diffusion models, recent works~\cite{prompt2prompt, g2023concept, g2025sliderspace, baumann2024continuous, chiu2025textsliderefficientplugandplay, dalva2024fluxspacedisentangledsemanticediting, yang2025controllablecontinuous} have extended continuous attribute control through per-attribute sliders or semantic embedding directions. Methods such as Concept Sliders~\cite{g2023concept} and \citet{baumann2024continuous} train per-attribute LoRAs or editing directions within the text embedding space to achieve smooth attribute manipulation. While all these approaches mark significant progress toward controllable generation, they face notable limitations: many require training a new LoRA or embedding direction per attribute, suffer from attribute entanglement, or degrade with multiple edits. They also primarily target \emph{text-to-image generation}, offering limited or indirect applicability to real-image editing. 

In contrast, our method introduces a novel and unified framework that generalizes slider-based continuous control to \emph{instruction-based image editing}. It eliminates the need for per-attribute retraining, supports multiple simultaneous edits, and remains robust across diverse editing scenarios and unseen attributes, while achieving significantly better performance on real-image editing tasks.

\section{SliderEdit: Continuous Image Editing}
\label{sec:method}

\begin{figure}[t]
    \centering
    \includegraphics[width=\linewidth]{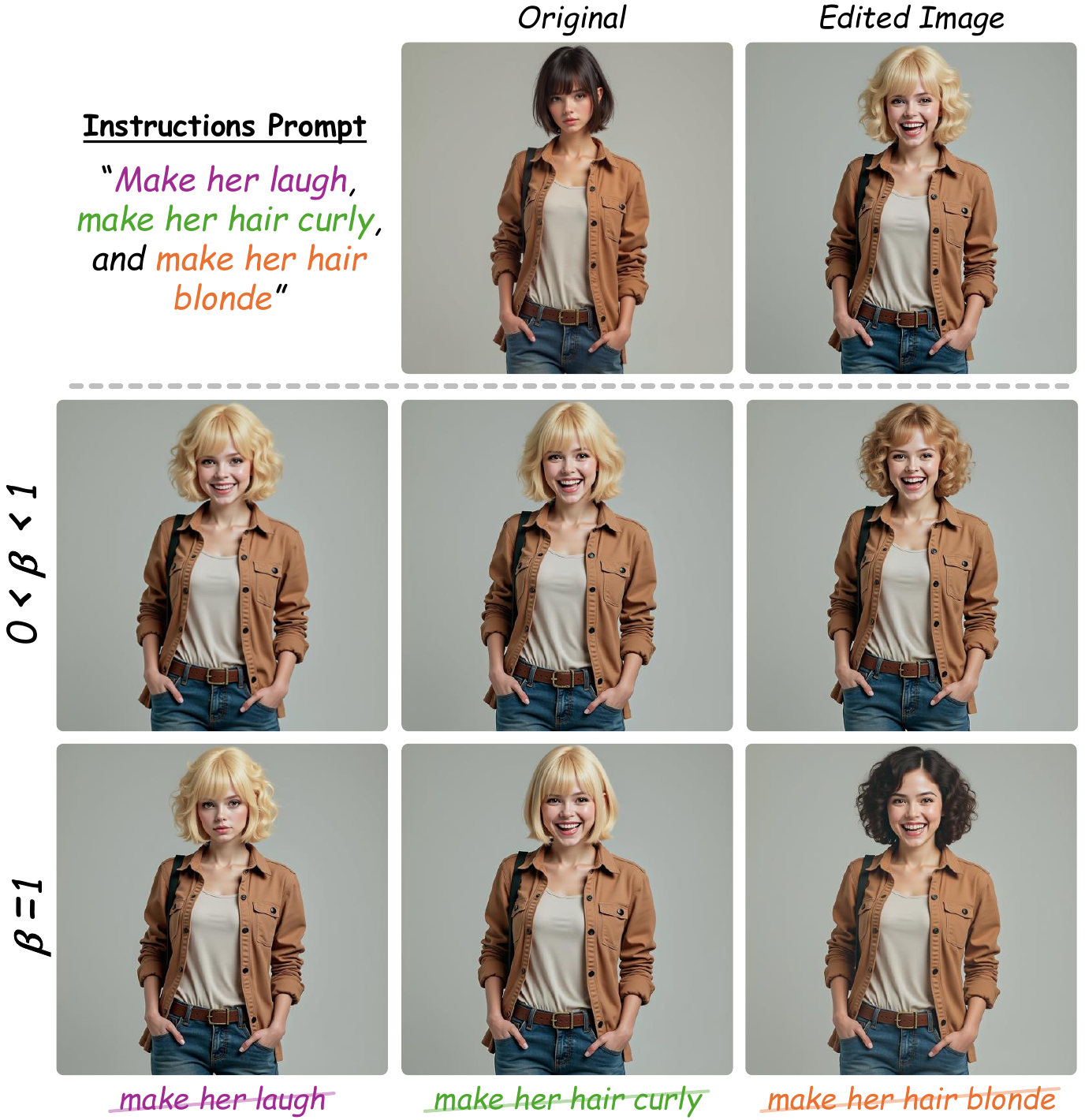}
    \vspace{-0.65cm}
    \caption{\textbf{Instruction-token embedding interpolation for strength control.} Interpolating between instruction and null-token embeddings produces intermediate edit strengths, demonstrating the potential for achieving fine-grained control through direct manipulation of intermediate instruction embeddings.}
    \label{fig:interpretability}
\end{figure}

In this section, we address the problem of enabling fine-grained control over individual editing instructions in a multi-instruction prompt for image editing. Formally, given a prompt $\mathcal{P} = \{ \mathcal{P}_1, ..., \mathcal{P}_K \}$, where each $\mathcal{P}_i$ denotes a distinct edit instruction (e.g., “make her laugh”, “make her hair curly”; see Fig.~\ref{fig:interpretability}, top row), our goal is to allow the user to modulate the strength of each instruction independently. To this end, we aim to associate each instruction $\mathcal{P}_i$ with a corresponding scaling factor $\beta_i \in [0, 1]$, allowing users to continuously control the strength of that specific edit—ranging from fully suppressing it ($\beta_i = 0$) to fully applying it ($\beta_i = 1$), or even exaggerating it when $\beta_i > 1$.

In Section~\ref{sec:method:mmdit_background}, we present the background of the MMDiT architecture and describe how text and image tokens are processed. Section~\ref{sec:method:interpretability} then examines how individual instructions $\mathcal{P}_i$ influence the generation process by tracing their effect through the model’s internal representations. This interpretability analysis provides key insights into where and how control can be applied. Building on this, Section~\ref{sec:method:main} introduces our method for modulating each instruction's strength via a continuous scaling mechanism, enabling fine-grained control over multi-instruction prompts.

\begin{figure*}[t]
    \centering
    \includegraphics[width=0.92\linewidth]{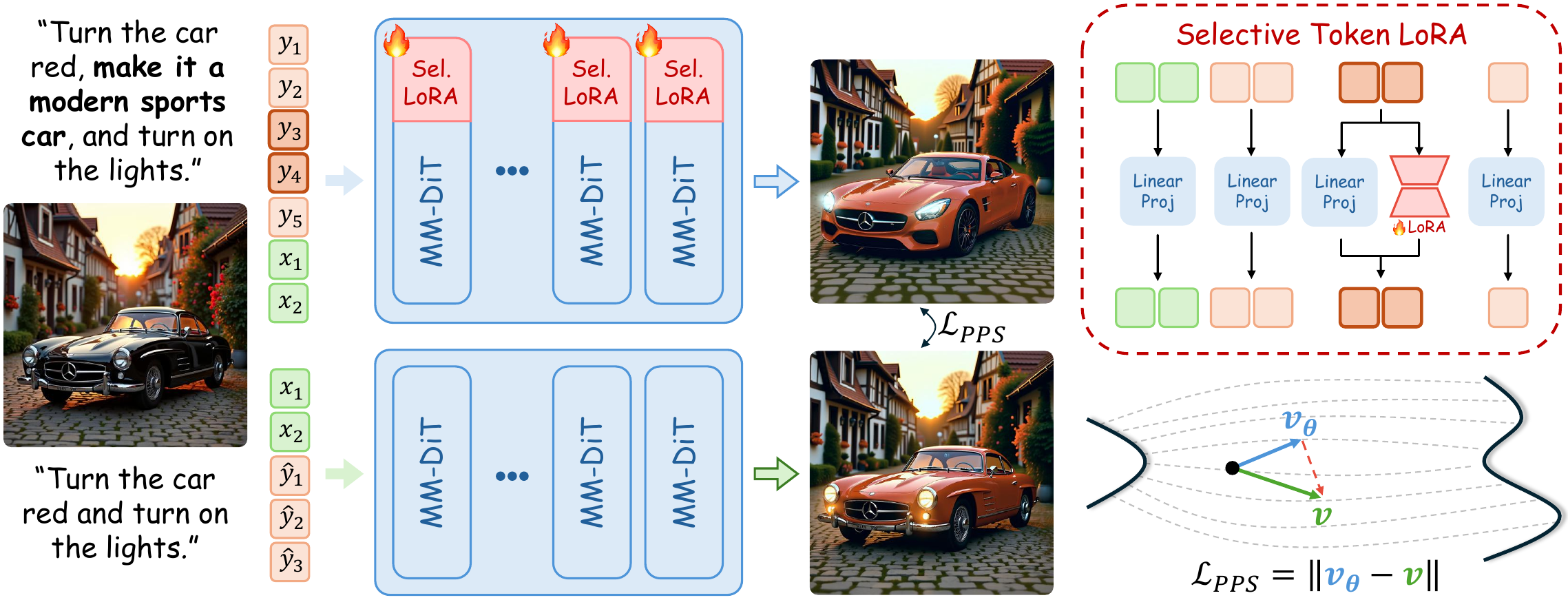}
    \vspace{-0.2cm}
    \caption{\textbf{Overview of the SliderEdit training pipeline.} Learnable low-rank matrices are applied to the intermediate token embeddings corresponding to the target edit instruction. These adapters are trained using the Partial Prompt Suppression (PPS) loss, which encourages the model to suppress or neutralize the visual effect of the selected instruction tokens.}
    \label{fig:method}
\end{figure*}

\subsection{MMDiT Architecture}
\label{sec:method:mmdit_background}
Recent image editing models such as FLUX-Kontext and Qwen-Image-Edit are built upon the MM-DiT architecture, which features a dual-branch structure: one for latent image tokens and one for text embeddings. Specifically, the latent image tokens ${x_1, \dots, x_N}$ represent both the noisy vectors in the VAE latent space and the encoded latents of the conditioning source image, while the text tokens ${y_1, \dots, y_T}$ are obtained by encoding the prompt $\mathcal{P}$ using a pretrained language model (e.g., T5). 

The prompt $\mathcal{P}$ is first tokenized into ${y_1', \dots, y_\tau'}$, then padded with the special \texttt{<pad>} token to reach a fixed length $T$: $\{y'_1, \ldots, y'_\tau, y'_{\tau+1} = y'_\texttt{<pad>}, \ldots, y'_T = y'_\texttt{<pad>} \}.$ These are passed through the T5 encoder to produce final text token embeddings ${y_1, \dots, y_T}$. The image tokens and text embeddings are then jointly processed by each MM-DiT block, where they interact through shared attention layers that enable cross-modal information exchange between visual and textual representations.


\subsection{Instruction-Level Interpretability Analysis}
\label{sec:method:interpretability}

Building on the previous section’s description of token interactions in MMDiT, we next investigate how individual instruction tokens affect generation. Specifically, we analyze the subset $\{y_u, y_{u+1}, \dots, y_{u'}\}$ corresponding to an edit instruction $\mathcal{P}_{\text{target}}$. These embeddings carry the semantic signal responsible for the target edit, and we test whether their influence is localized or diffused through the network via targeted interventions.

Specifically, within each attention block at layer $\ell$, we intervene on the target instruction embeddings $\{y^\ell_u,y^\ell_{u+1},...,y^\ell_{u'}\}$, which represent the target instruction tokens input to that block (i.e., the embeddings after processing by the preceding layers). We linearly interpolate them with the padding token embedding $y^\ell_{\texttt{<pad>}}$:
\[
y^\ell_j \leftarrow (1-\beta) \cdot y^\ell_j + \beta \cdot y^\ell_{\texttt{<pad>}}, \quad \text{for } j \in \{u, \ldots, u'\}.
\]
The interpolation coefficient $\beta \in [0, 1]$ determines how much of the instruction’s information is preserved. Setting $\beta = 1$ effectively removes the instruction by replacing its embeddings with that of the padding token (i.e., no information), while $\beta = 0$ leaves the instruction fully intact.

Figure~\ref{fig:interpretability} illustrates the resulting generations. The bottom row corresponds to $\beta = 1$, where the edit is entirely removed, while the middle row shows an intermediate, manually chosen $\beta$ that partially applies the edit. These results demonstrate that the intermediate token embeddings corresponding to $\mathcal{P}_{\text{target}}$ are highly localized and show strong potential for achieving fine-grained control by directly manipulating their embeddings. While this analysis shows potential for controlling edits via simple embedding interpolation, this approach provides only limited and discontinuous modulation. To achieve stronger and smoother control, we propose a robust method in the next section.

\subsection{Fine-Grained Control of Edit Instructions}
\label{sec:method:main}


Building on our interpretability findings, we introduce a mechanism that enables continuous and independent control over each edit instruction in a multi-instruction prompt. 

Given an input image $X_\text{orig}$ and a prompt $\mathcal{P} = \{ \mathcal{P}_1, ..., \mathcal{P}_K \}$ containing $K$ edit instructions, a base image editing model produces an edited output $X_\text{editted}^{\mathcal{P}_1, ..., \mathcal{P}_K}$ where all edits are applied simultaneously. Our objective is to learn a flexible adapter $M_\theta(\mathcal{P}_i)$ capable of suppressing or modulating a specific instruction $\mathcal{P}_i$ within $\mathcal{P}$. When this adapter is activated, the model should generate $X_\text{editted}^{\mathcal{P}_1, ..., \mathcal{P}_{i-1}, \mathcal{P}_{i+1} ,...,\mathcal{P}_K}$, effectively removing the influence of $\mathcal{P}_i$ while keeping other edits intact.

\vspace{2pt}
\textbf{Partial Prompt Suppression Loss.}  To train $M_\theta$, we propose the Partial Prompt Suppression (PPS) objective. Using the frozen base model $\epsilon(Z, X, P)$, where $Z$ denotes the noisy latents, $X$ the original image latents, and $P$ the text prompt, we first perform a forward pass with the prompt excluding the $i$-th instruction $\mathcal{P}_i$. We then require that the adapted model $\epsilon_{M_\theta(\mathcal{P}_i)}$, when given the full prompt, produces an equivalent denoising direction:
\[
\mathcal{L}_\texttt{PPS} = \|\epsilon_{M_\theta(\mathcal{P}_i)}(Z, X_\text{orig}, \mathcal{P}) -\epsilon(Z, X_\text{orig}, \mathcal{P}-\{\mathcal{P}_i\}) \|
\]
Intuitively, this objective teaches the adapter to neutralize the representation of the tokens corresponding to $\mathcal{P}_i$ throughout the model so that their visual effect disappears.
In addition to PPS, we introduce a simplified variant, \emph{Simplified Partial Prompt Suppression (SPPS)}. SPPS treats each edit prompt as a single instruction (i.e., $\mathcal{P} = {\mathcal{P}_1}$) and applies the same suppression objective directly to $\mathcal{P}_1$ (See Figure \ref{fig:method_simple}). Despite its simplicity, SPPS yields highly robust and generalizable adapters, even for multi-instruction editing scenarios. Algorithm \ref{alg:pps} outlines the overall training procedure of SliderEdit. Additional details on SPPS and its comparison with PPS are provided in Appendix \ref{sec:app:spps}.

\begin{algorithm}[t]
\caption{Training SliderEdit}
\label{alg:pps}
\begin{algorithmic}[1]
\Require $\epsilon(Z, X, P)$: Image Editing Model, $M_\theta$: Trainable Adapter, $\{(X^{(i)}_\text{orig}, \mathcal{P}^{(i)})\}$: Dataset
\For{each training step}
    \State $X_\text{orig}, \mathcal{P}=\{\mathcal{P}_1,\ldots,\mathcal{P}_K\}$ $\leftarrow$ Sample a Data
    \State $\varepsilon \sim \mathcal{N}(0,I), t \sim \mathcal{U}[0,1]$
    \If{use $\mathcal{L}_\text{SPPS}$}
        \State $\mathcal{P} = \{\mathcal{P}'_1\}$ \textcolor{gray}{\Comment{{\scriptsize Consider whole as a single-instruction prompt}}}
    \EndIf
    \State $Z \leftarrow (1-t)\varepsilon + tX_\text{orig}$
    \State $\mathcal{P}_i \leftarrow$ Random target instruction from $\mathcal{P}$ to suppress
    \State $v^\star \leftarrow \epsilon(Z, X_\text{orig}, \mathcal{P}\setminus\{\mathcal{P}_i\})$
    \State $\hat{v} \leftarrow \epsilon_{M_\theta(\mathcal{P}_i)}(Z, X_\text{orig}, \mathcal{P})$
    \State $\mathcal{L}_\texttt{PPS} = \|\hat{v} - v^\star\|^2$
    \State Update $\theta$ via gradient descent on $\mathcal{L}_\texttt{PPS}$
\EndFor
\vspace{2pt}
\end{algorithmic}
\end{algorithm}

\makeatletter
\newcommand{\INPUT}{\item[\textbf{Input:}]} 
\makeatother
\begin{algorithm}[t]
\caption{$M_\theta^\ell$ (STLoRA / GSTLoRA)}
\label{alg:stlora_and_gstlora}
\begin{algorithmic}[1]

\Require $W^\ell$: Base Linear Projection, $\{A^\ell, B^\ell\}$: Low-Rank Matrices, \texttt{mode} $\in \{\texttt{STLoRA}, \texttt{GSTLoRA}\}$
\INPUT $\{x_1,\ldots,x_N\}$: Image Tokens, $\{y_1,\ldots,y_T\}$: Text Tokens, $\mathcal{P}_i$: Target Instruction
\vspace{4pt}
\State $\Delta W^\ell = B^\ell A^\ell$
\If{\texttt{mode} $= \texttt{GSTLoRA}$}
    \State $y_i \leftarrow (W+\Delta W)y_i \quad\quad\; \forall y_i \in \{y_1,\dots,y_T\}$
    \State $x_i \leftarrow (W+\Delta W)x_i\quad\quad\,\forall x_i \in \{x_1,\dots,x_N\}$
\ElsIf{\texttt{mode} $= \texttt{STLoRA}$}
    \State $\mathcal{T} \leftarrow \text{TokenIndices}(\mathcal{P}_i)$ \textcolor{gray}{\Comment{{\scriptsize Indices in $\{1,\dots,T\} \mapsto\mathcal{P}_i$}}}
    \State $y_i \leftarrow (W+\Delta W)y_i \quad\quad\; \forall y_i \in \mathcal{T}$
    \State $y_i \leftarrow Wy_i \quad\quad\qquad\qquad\, \forall y_i \in \{y_1,\dots,y_T\} \setminus \mathcal{T}$
    \State $x_i \leftarrow Wx_i \qquad\qquad\qquad \forall x_i \in \{x_1,\dots,x_N\}$
\EndIf
\State \Return $\{x_1,\ldots,x_N\}, \{y_1,\ldots,y_T\}$
\vspace{2pt}
\end{algorithmic}
\end{algorithm}

\vspace{2pt}
\textbf{Selective Token LoRA.}
We instantiate $M_\theta$ as a Selective Token LoRA (STLoRA)—a lightweight, token-aware adapter.
STLoRA learns low-rank updates for selected linear projections in the model but applies them \textit{only} to the embeddings of target tokens corresponding to the suppressed instruction $\mathcal{P}_i$. Formally, consider a linear projection at layer $\ell$ where tokens $z$ (either image or text) are transformed as $z' = W^\ell z$. STLoRA introduces trainable low-rank matrices $A^\ell$ and $B^\ell$ with $\Delta W^\ell = B^\ell A^\ell$, updating only the selected target tokens:
\[
z'_\text{target} = (W^\ell + \Delta W^\ell) z_\text{target}, \quad
z'_\text{others} = W^\ell z_\text{others}.
\]
This selectivity ensures the adapter modifies only target token embeddings. Figure~\ref{fig:method} illustrates the \emph{SliderEdit} training pipeline, and Figure~\ref{fig:method_simple} shows the SPPS variant.

\vspace{2pt}
\textbf{Continuous Control via Scaling STLoRA}
Once trained, the LoRA adapter naturally supports continuous control through its scaling parameter \cite{hu2022lora, g2023concept, shah2024ziplora}.
We denote $M_\theta^\alpha$ as the adapter with scaled updates $\alpha \Delta W_\ell$ for each layer.
By varying $\alpha$ within a predefined range $[\alpha_\text{min}, \alpha_\text{max}]$, we obtain a smooth continuum of effects—from complete suppression ($\alpha = 1$) to full application ($\alpha = 0$), and even exaggerated edits for $\alpha < 0$. Note that the scaling parameter $\alpha_i$ follows an inverse range compared to $\beta_i$ defined earlier. The two scales can be related through $\alpha = 1 - \beta$.

\vspace{2pt}
\textbf{Globally Selective Token LoRA}
While STLoRA effectively handles both single- and multi-instruction prompts by selectively modulating tokens corresponding to each instruction $\mathcal{P}_i$, we introduce Globally Selective Token LoRA (GSTLoRA) for the single-instruction setting. In this variant, all token embeddings (both text and image) are included in the adaptation, allowing LoRA updates to be applied globally across the representation space. This design provides stronger control and often yields higher-fidelity edits when manipulating a single instruction, as the update can leverage global context rather than being limited to a subset of intermediate text token embeddings. Algorithm~\ref{alg:stlora_and_gstlora} outlines the operation of STLoRA and GSTLoRA adapters.

\begin{figure*}[t]
    \centering
    \includegraphics[width=0.83\linewidth]{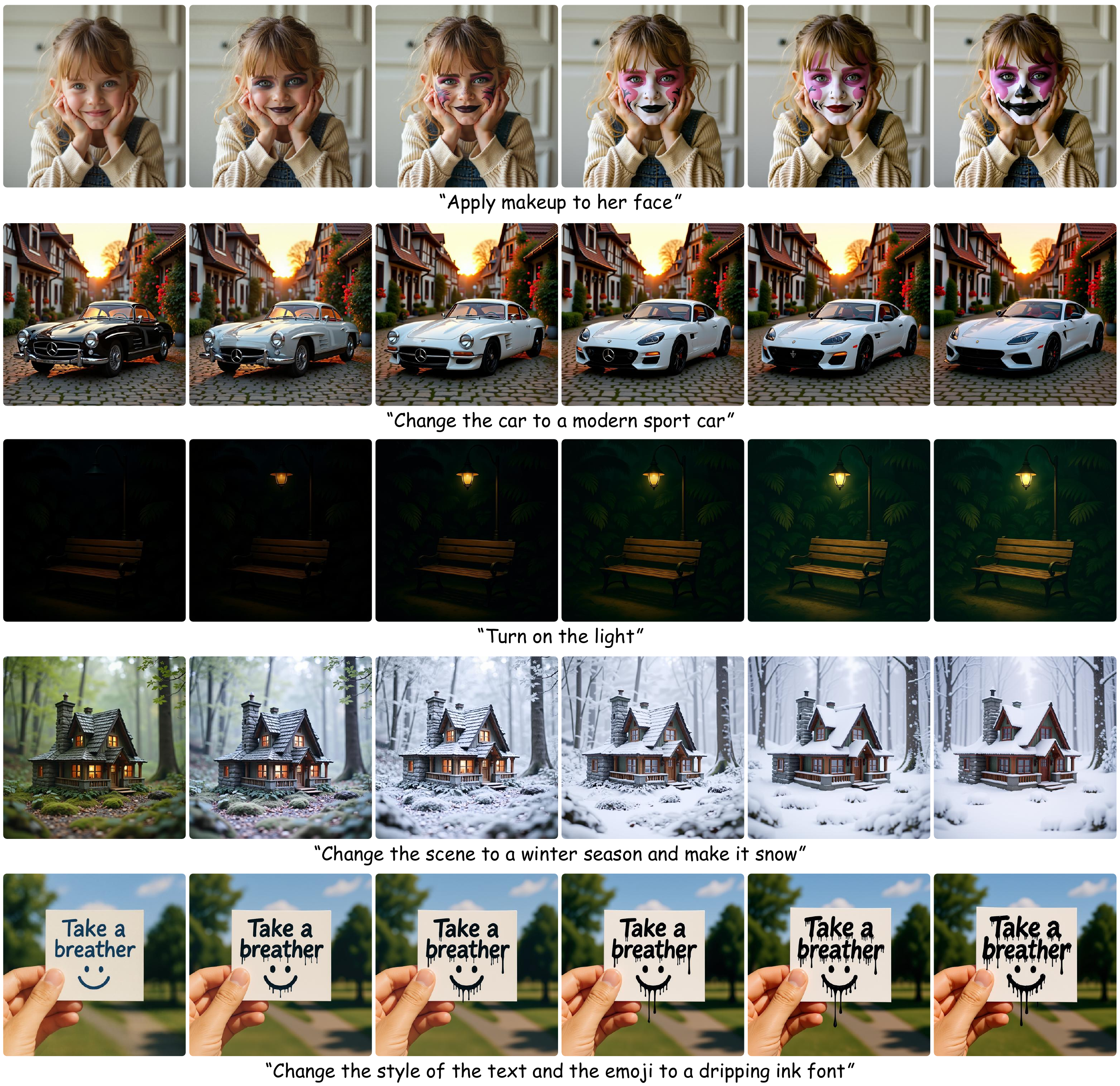}
    \vspace{-0.3cm}
    \caption{\textbf{Qualitative Samples of GSTLoRA.} Demonstrates smooth, continuous control over the strength of both local and global edits.}
    \label{fig:gstlora_qualitative}
\end{figure*}

\section{Experiments}
\label{sec:experiments}

We conduct comprehensive quantitative and qualitative evaluations of SliderEdit, showing that it performs robustly across a wide range of instruction edits. In addition, we compare various baselines and SliderEdit variants, demonstrating that our method achieves superior results, offering continuous and precise control over edits.

\subsection{Implementation details}
\label{sec:implementation_details}
\looseness=-1
We use FLUX-Kontext and Qwen-Image-Edit as our base models. All models are trained with the $\ell_{\text{SPPS}}$ loss for simplicity and generalization, while $\mathcal{L}_{\text{PPS}}$ provides stronger multi-instruction control for STLoRA (see Appendix~\ref{sec:app:spps}). We set the LoRA rank to 16, keeping the adapters lightweight and efficient.
Training uses a small subset (1k–8k samples) of the GPT-Image-Edit dataset~\cite{wang2025gpt}.
Both STLoRA models are trained for 1,000 iterations, converging around 400 but extended for consistency.
GSTLoRA on FLUX-Kontext is trained for 300 iterations. Overall, the training process is computationally very lightweight and data-efficient.
Further details are provided in Appendix~\ref{sec:app:implementation_details}.

\subsection{Qualitative Results}
\label{sec:qual_results}
In this section, we qualitatively evaluate the results of SliderEdit variants, demonstrating their effectiveness across diverse scenarios and editing capabilities.

Figure~\ref{fig:gstlora_qualitative} presents qualitative examples of GSTLoRA applied to the FLUX-Kontext model. As shown, our method produces smooth and continuous edit trajectories, enabling fine-grained control over the strength of edits. It effectively handles both \emph{local edits} (e.g., adding makeup or modifying a car's age) and \emph{global edits} (e.g., changing the season or adjusting scene lighting). Additional examples, such as edits involving camera view or angle changes, as well as applications in text editing and face editing, are provided in Figures~\ref{fig:app:gstlora_qualitative},~\ref{fig:gstlora_text_slider_appendix},~\ref{fig:app:gstlora_qualitative_face_1},~and~\ref{fig:app:gstlora_qualitative_face_2} in the Appendix.

Figures~\ref{fig:2d_stlora_girl}~and~\ref{fig:app:stlora_2d_qualitative} illustrate qualitative results of STLoRA for instructions containing two edit directions. The resulting 2D intermediate space exhibits smooth and continuous variations, allowing users to precisely control edit strengths along each direction to obtain desired outputs. Figures \ref{fig:teaser} and \ref{fig:app:stlora_3d_qualitative} provide additional cases with three edit directions. 

\begin{figure}[t]
    \centering
    \includegraphics[width=\linewidth]{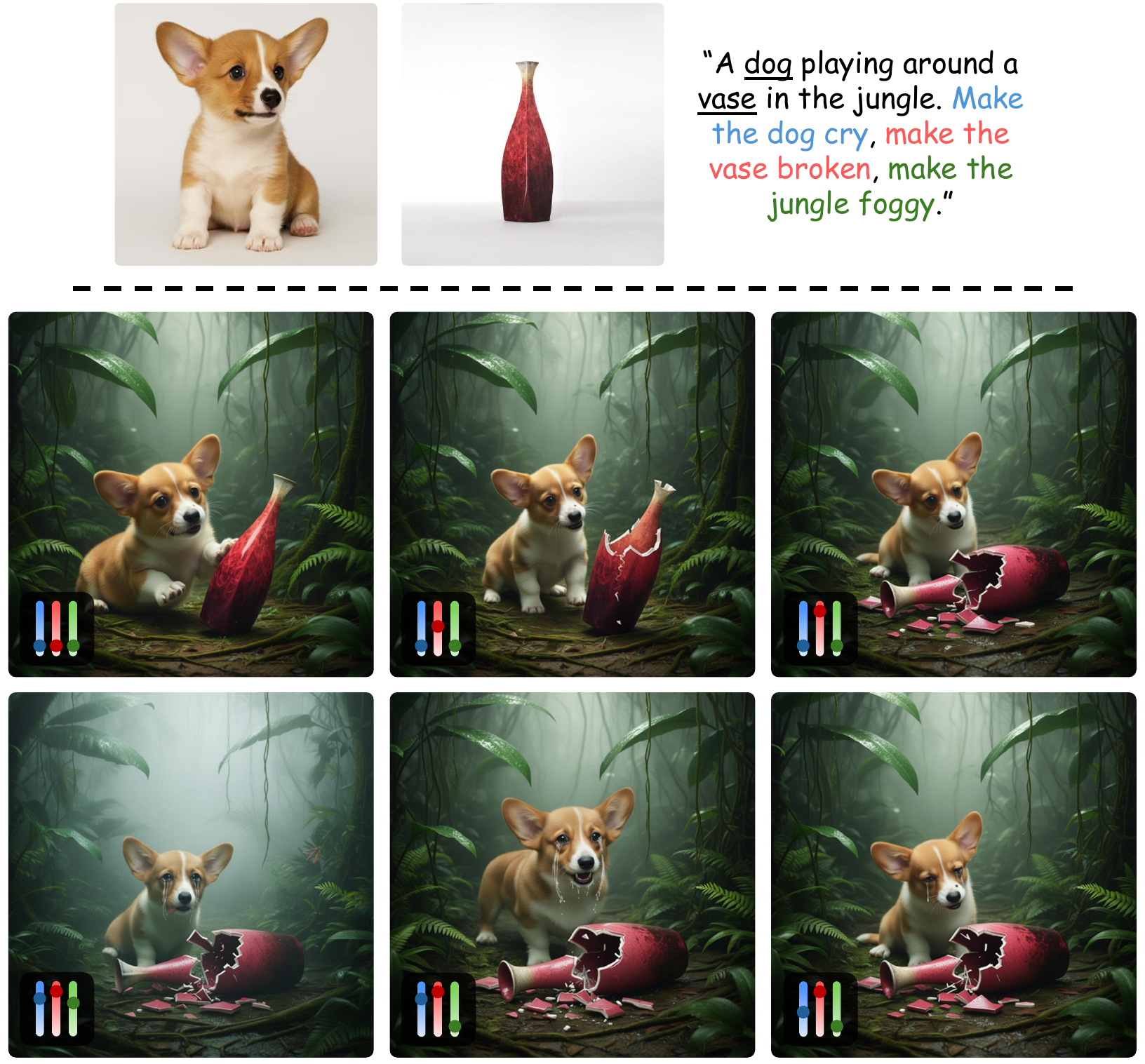}
    \vspace{-0.65cm}
    \caption{\textbf{Controllable zero-shot multi-subject personalization with STLoRA.} STLoRA enables smooth adjustment of each instruction’s strength to generate coherent, evolving image sequences, supporting story-like visual editing. (Best viewed from top-left to top-right, then bottom-right to bottom-left)}
    \label{fig:multi_subject_personalization}
\end{figure}

To further explore the versatility of SliderEdit, we examine its performance on advanced tasks supported by state-of-the-art editing models. One such task is \emph{zero-shot personalization}. Figure~\ref{fig:multi_subject_personalization} shows an example where STLoRA is integrated with Qwen-Image-Edit to perform multi-subject personalization, followed by instruction-based scene editing. Our approach provides users with flexible, fine-grained control—by adjusting sliders, one can generate a coherent series of images that naturally evolve, resembling a narrative. This demonstrates the potential of SliderEdit as a powerful tool for storytelling and creative content generation.

\subsection{Quantitative Results}
\label{sec:quan_results}

In this section, we quantitatively evaluate the performance of STLoRA and GSTLoRA against multiple baselines. We assess their ability to achieve continuous, extrapolative, and disentangled control through quantitative metrics, providing an objective analysis of the smoothness and independence of the edit trajectories generated by each method.

\subsubsection{Evaluation Set} 
For quantitative evaluation, we construct a facial editing benchmark with $N$ subjects of diverse genders, ages, and ethnicities, and define $M$ edit directions (e.g., \textit{"make the hair curly"}, \textit{"make the hair long"}). Original images are chosen so that target attributes are absent (e.g., straight, short hair). We evaluate each model under editing configurations containing $\gamma$ instructions, sampling $\gamma$ instructions from the $M$ available to form $\binom{M}{\gamma}$ prompts. For each instruction, the edit strength $\alpha$ varies within $\left[\alpha_{\text{min}}, \alpha_{\text{max}}\right]$ across $\delta$ steps, yielding a $\gamma$-dimensional edit space of $\delta^{\gamma}$ images per prompt. This structured space enables quantitative analysis of \emph{continuity}, \emph{extrapolation}, and \emph{disentanglement}.



\subsubsection{Metrics}
We employ several quantitative metrics to evaluate different aspects of the editing behavior, including \emph{continuity}, \emph{extrapolation}, and \emph{disentanglement}.  
For each instruction edit, the model generates a sequence of images 
at varying edit strengths,
which are then analyzed using the following metrics.
To measure how strongly each edit is reflected in the generated image, we use vision-language models.
For each instruction (e.g., “make the person laugh”), we define a corresponding descriptive prompt (e.g., “a person smiling”) and compute image–text similarity in the embedding space of VLMs such as \emph{CLIP}~\cite{radford2021learning}, \emph{SigLIP}~\cite{zhai2023sigmoid}, and \emph{BLIP}~\cite{li2022blip}.
This score measures how well the edit is executed.


\begin{figure}[t]
    \centering
    \includegraphics[width=0.90\linewidth]{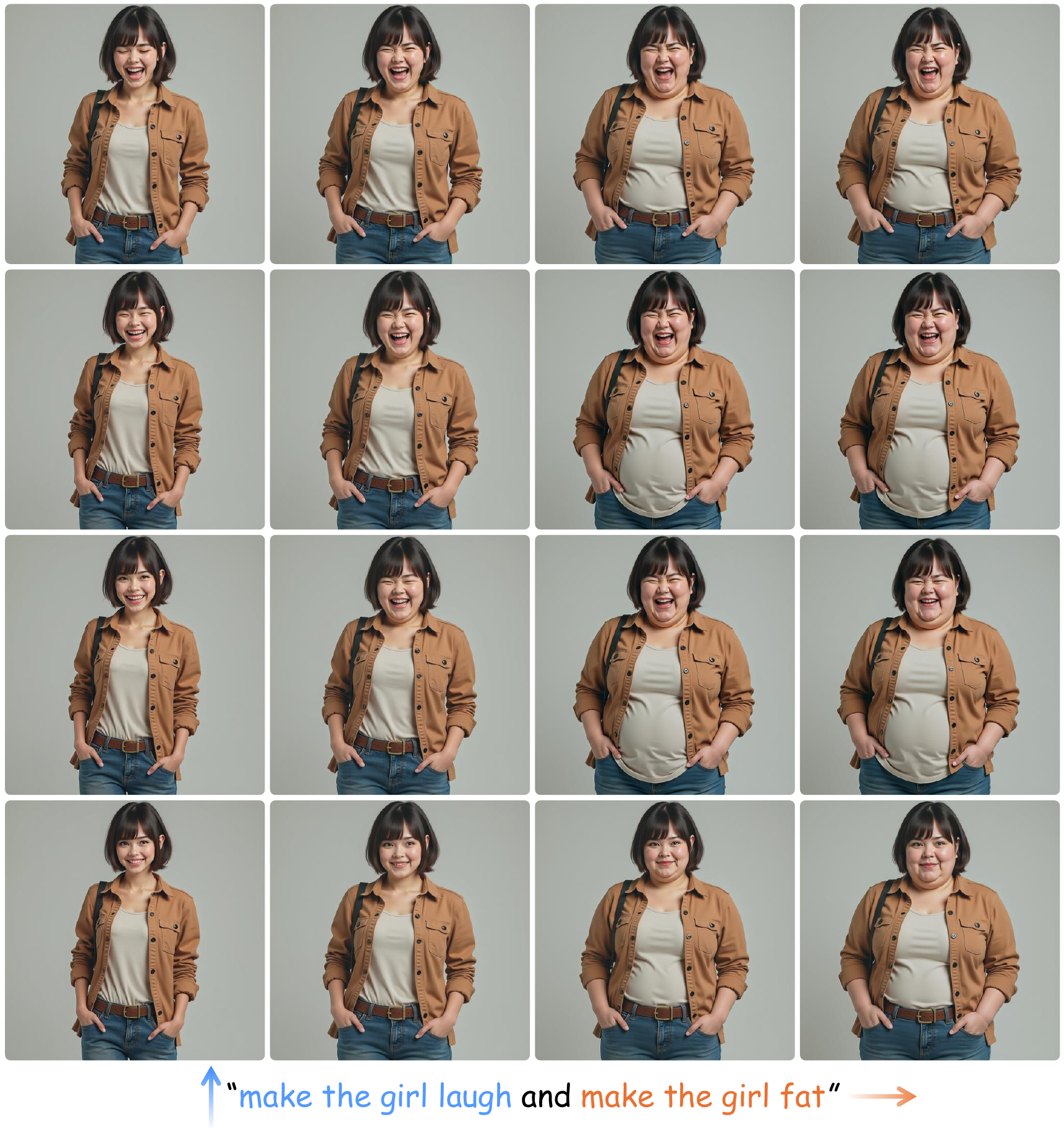}
    \vspace{-0.35cm}
    \caption{\textbf{Qualitative results of STLoRA on 2-instruction edit.} The 2D grid shows smooth, continuous transitions, allowing precise and disentangled control over each instruction's strength.}
    \label{fig:2d_stlora_girl}
\end{figure}

\begin{table*}[t]
\centering
\small
\renewcommand{\arraystretch}{0.5}
\setlength{\tabcolsep}{5pt}
\resizebox{\linewidth}{!}{\begin{tabular}{lcccccccccc}
\toprule
& \multicolumn{3}{c}{\bf Extrapolation $\uparrow$} & \multicolumn{3}{c}{\bf Continuity $\uparrow$} & \multicolumn{4}{c}{\bf Disentanglement $\downarrow$} \\
\cmidrule(lr){2-4} \cmidrule(lr){5-7} \cmidrule(lr){8-11}
  \multirow{-2}{*}{\bf Model}  & \bf CLIP & \bf SigLIP & \bf BLIP & \bf CLIP & \bf SigLIP & \bf BLIP & $ \text{\bf LPIPS}_\text{\bf alex}$ & $\text{\bf LPIPS}_\text{\bf vgg}$ & \bf DiNO  & \bf ID  \\
\midrule
Concept Slider \cite{g2023concept} & 0.2581  & 0.0666 & 0.4680 & 0.1803  & 0.2071  & 0.1719   & 0.2174 & 0.2495   & 0.2140   & \underline{0.7091} \\
Cont. Attr. Control \cite{baumann2024continuous}  & 0.2380  & 0.0484 & 0.2082  & 0.1891  &  0.2167 & 0.1114   & 0.1973 &  0.2485  & 0.1852 & \underline{0.5519}  \\
  Implicit CFG & 0.2607 & 0.0667 & 0.4658 & 0.1547 & 0.1906 & 0.1513  & 0.2149 & 0.2611  & 0.1405 & 0.2748 \\
  Explicit CFG \cite{ho2022classifier} & \cellcolor{green!15}0.2731  & \cellcolor{green!15}0.0739   & \cellcolor{green!15}0.5717    & 0.1993    & 0.2263    & 0.1803    &  0.2465    & 0.3085    & 0.1718    & 0.3415    \\
\midrule
  \bf $\text{SliderEdit}_{\text{STLoRA}}$ & 0.2632  & 0.0669  & 0.4762   & \cellcolor{yellow!15}0.2538    & \cellcolor{yellow!15}0.2495   & \cellcolor{yellow!15}0.1830   &  \cellcolor{yellow!15} 0.1902   & \cellcolor{yellow!15} 0.2345   & \cellcolor{green!15}\bf 0.1071   & \cellcolor{green!15}\bf 0.2550   \\
  \bf $\text{SliderEdit}_{\text{GSTLoRA}}$  & \cellcolor{yellow!15}0.2648 &  \cellcolor{yellow!15}0.0673  & \cellcolor{yellow!15}0.4862  & \cellcolor{green!15}\textbf{0.2998}   & \cellcolor{green!15}\textbf{0.3062}  & \cellcolor{green!15}\textbf{0.2227}  & \cellcolor{green!15}\bf 0.1868   & \cellcolor{green!15}\bf 0.2330   & \cellcolor{yellow!15}0.1168 & \cellcolor{yellow!15}0.2675    \\
\bottomrule
\end{tabular}}
\vspace{-0.30cm}
\caption{\textbf{Quantitative results for single-instruction edits ($\gamma = 1$).} SliderEdit yields smoother trajectories and better identity preservation.
}
\label{tab:1d_comparison}
\end{table*}

\textbf{Extrapolation.}
Extrapolation measures the model’s ability to apply edits beyond the standard range, which is particularly useful when amplifying attributes such as facial expressions.
We define the extrapolation score as the maximum VLM's similarity value, 
which indicates the strongest expression of the target attribute achieved by the model.

\begin{table}[t]
\vspace{-0.4cm}
\centering
\renewcommand{\arraystretch}{0.8}
\setlength{\tabcolsep}{4pt}
\resizebox{0.98\linewidth}{!}{\begin{tabular}{llccc}
\toprule
$\gamma$ & \textbf{Model} & \textbf{$\text{Extrap.}_\text{Avg}$ $\uparrow$} & \textbf{$\text{Cont.}_\text{Avg}$ $\uparrow$} & \textbf{$\text{Dis.}_\text{Avg}$ $\downarrow$} \\
\midrule
\multirow{2}{*}{1} 
& FLUX$_{\text{Kontext}}$ & 0.2630 & 0.2709 & 0.1912 \\
& Qwen$_{\text{Image-Edit}}$ & 0.3214 & 0.2319 & 0.2187 \\
\midrule
\multirow{2}{*}{2} 
& FLUX$_{\text{Kontext}}$ & 0.2776 & 0.2409 & 0.2509 \\
& Qwen$_{\text{Image-Edit}}$ & 0.3160 & 0.2813 & 0.3088 \\
\midrule
\multirow{2}{*}{3} 
& FLUX$_{\text{Kontext}}$ & 0.2970 & 0.3691 & 0.2762 \\
& Qwen$_{\text{Image-Edit}}$ & 0.3417 & 0.4345 & 0.3630 \\
\bottomrule
\end{tabular}}
    \vspace{-0.25cm}
\caption{\textbf{Quantitative results for multi-instruction edits.} Both models show comparable performance in continuity. FLUX better preserves identity, while Qwen performs better in extrapolation.}

\label{tab:2d_quantitative_comparison}
\end{table}

\textbf{Continuity.} 
Given similarity scores ${s_1, \ldots, s_\delta}$ for increasing $\alpha$ values, we expect them to vary smoothly and uniformly between $\min(s_i)$ and $\max(s_i)$.
We quantify this using a chi-squared statistic comparing the observed and expected counts of $s_i$ across bins, where higher $({\chi^2_{\text{agg}}}/{\text{dof}})^{-1}$ indicates smoother edit trajectories.
For 2D and 3D edit spaces, we apply the same test to assess the uniformity across the grids. For more details, refer to Appendix \ref{sec:app:metrics}

\textbf{Disentanglement.}
We evaluate how well the model applies an edit without altering unrelated factors such as identity or background. \emph{Identity preservation} is measured via cosine distance in the \emph{ArcFace}~\cite{deng2019arcface} embedding space, where lower values indicate better consistency.
We further compute perceptual distances between edited and original images using \emph{LPIPS}~\cite{zhang2018unreasonable} (AlexNet~\cite{NIPS2012_c399862d}, VGG~\cite{simonyan2014very}) and \emph{DINOv2}~\cite{caron2021emerging, oquab2023dinov2}, capturing both low-level perceptual and high-level semantic changes to assess overall disentanglement.


\subsubsection{Baselines}
We consider different baselines depending on the number of edit instructions $\gamma$ used in the prompt. 

For the case of a single-instruction setting ($\gamma = 1$), we compare \emph{GSTLoRA} (Ours) and \emph{STLoRA} (Ours) with \emph{Explicit CFG} and \emph{Implicit CFG}, all implemented on top of the \emph{FLUX-Kontext} model, as well as prior methods Concept-Slider \cite{g2023concept} and Continuous Attribute Control \cite{baumann2024continuous}.  FLUX-Kontext is a guidance-distilled model that internally approximates the effect of classifier-free guidance, allowing implicit control over edit strength but offering limited flexibility.
To enable explicit control, we reconfigure it to perform guidance externally during inference.
Concept-Slider and Continuous Attribute Control provide fine-grained attribute manipulation but rely on inversion techniques~\cite{mokady2023null, garibi2024renoise}, making them less effective for direct image editing.

For multi-instruction edits ($\gamma > 1$), Explicit CFG, Implicit CFG, and GSTLoRA cannot independently control each edit direction, whereas \emph{STLoRA} enables disentangled, per-instruction control.
Since Concept-Slider and Continuous Attribute Control already perform poorly in single-instruction settings, we omit them from this scenario.
We evaluate STLoRA on both \emph{FLUX-Kontext} and \emph{Qwen-Image-Edit}. For more details on baselines, refer to Appendix~\ref{sec:app:baselines}.

\subsubsection{Results}

Table~\ref{tab:1d_comparison} presents the quantitative comparison for single-instruction prompts ($\gamma = 1$) using $\delta = 15$ across different baselines and metrics, including \emph{continuity}, \emph{extrapolation}, and \emph{disentanglement}. For fair comparison, continuity is calculated based on normalized scores across all methods. As shown, \emph{GSTLoRA} achieves the highest continuity while maintaining strong disentanglement and satisfactory extrapolation performance. Notably, although one might expect \emph{Explicit CFG} to perform comparably, both \emph{STLoRA} and \emph{GSTLoRA} significantly outperform it, demonstrating superior smoothness and control in edit strength. 

\begin{figure}[t]
\vspace{-0.55cm}
    \centering
    \includegraphics[width=\linewidth]{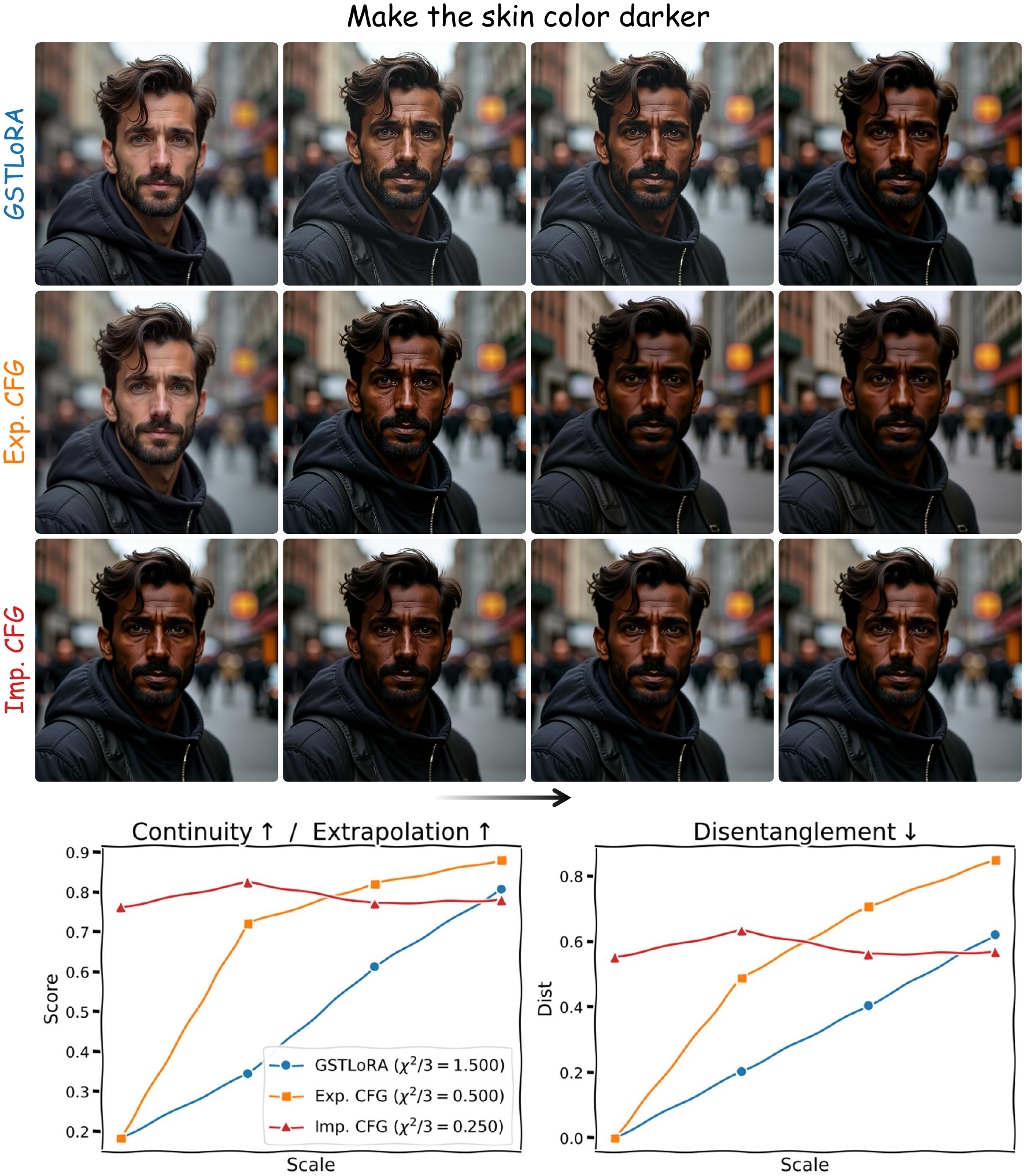}
    \vspace{-0.7cm}
    \caption{\textbf{Qualitative and quantitative comparison of GSTLoRA with CFG baselines.} GSTLoRA shows smooth edit trajectories with gradual similarity changes, unlike Implicit and Explicit CFG, which exhibit abrupt transitions and greater identity drift.}
    \label{fig:eval_qualitative_quantitative_sample}
\end{figure}

Figure~\ref{fig:eval_qualitative_quantitative_sample} provides qualitative and quantitative examples for a representative case. GSTLoRA produces a remarkably smooth and continuous edit trajectory, in contrast to \emph{Implicit} and \emph{Explicit CFG}, which exhibit abrupt transitions and inconsistent edit intensities. This behavior is well captured by our quantitative metrics—on the left, the aggregated average similarity score across normalized VLM metrics increases gradually for GSTLoRA, whereas both CFG variants show sudden jumps. In terms of disentanglement, GSTLoRA also achieves lower identity drift and more stable visual consistency compared to the other methods. Refer to Appendix \ref{sec:app:quantitative} for more comparison and other baselines.

Table~\ref{tab:2d_quantitative_comparison} presents quantitative results for multi-instruction prompts with $\gamma \in {1, 2, 3}$ and using $\delta = 7$, comparing \textit{FLUX-Kontext} and \textit{Qwen-Image-Edit}. Both models perform strongly: for single-instruction edits, STLoRA on \textit{FLUX} achieves better performance in terms of continuity, whereas in the two- and three-instruction settings, \textit{Qwen} demonstrates stronger results. 
Moreover, FLUX better preserves identity and disentanglement, whereas Qwen performs better in extrapolation.
However, as observed across all configurations, there consistently exists a trade-off between continuity, extrapolation, and disentanglement.

\section{Conclusion}
We introduced \emph{SliderEdit}, a unified framework for continuous, fine-grained instruction control in instruction-based image editing models. By training lightweight low-rank adapters with a novel loss to disentangle and modulate instruction effects, SliderEdit enables smooth, interpretable control over edit strength. Integrated with state-of-the-art models like \textit{FLUX-Kontext} and \textit{Qwen-Image-Edit}, it achieves superior controllability, visual coherence, and flexibility, laying the foundation for interactive, instruction-driven editing with continuous and compositional control.

\iftoggle{cvprfinal}{
\section*{Acknowledgement}
This project was supported in part by a grant from an NSF CAREER AWARD 1942230, the ONR PECASE grant N00014-25-1-2378, ARO’s Early Career Program Award 310902-00001, Army Grant No. W911NF2120076, the NSF award CCF2212458, NSF Award No. 2229885 (NSF Institute for Trustworthy AI in Law and Society, TRAILS), a MURI grant 14262683, DARPA AIQ DARPA AIQ grant HR00112590066  and an award from meta 314593-00001.
}{}

{
    \small
    \bibliographystyle{ieeenat_fullname}
    \bibliography{main}
}

\clearpage
\maketitlesupplementary

\section{Related Works}
\subsection{Diffusion Models and Flow Matching}
\label{app:related_works_diffusion_and_flow}

\begin{figure}[t]
    \centering
    \includegraphics[width=\linewidth]{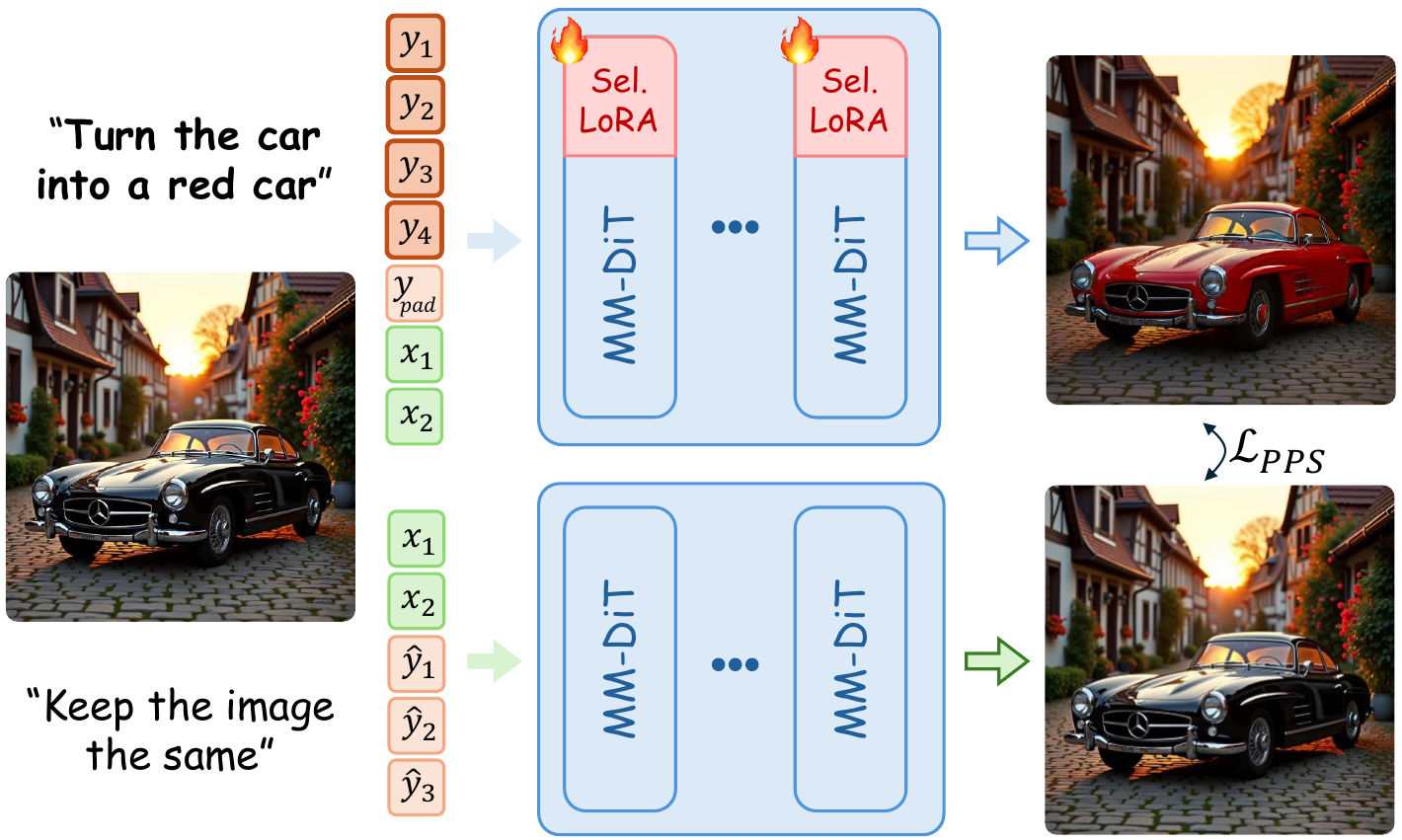}
    \caption{\textbf{Simplified Partial Prompt Suppression (SPPS).}
SPPS applies the same suppression objective as PPS but treats the entire edit prompt as a single instruction. During training, a second (bottom-row) forward pass is performed to obtain a neutralized image—either using an empty prompt (“”) or a neutral textual instruction (e.g., “keep the image the same”). This simple formulation effectively teaches the adapter to suppress undesired edit effects and generalizes well to multi-instruction editing scenarios.}
    \label{fig:method_simple}
\end{figure}

\begin{figure}[t]
    \centering
    \includegraphics[width=\linewidth]{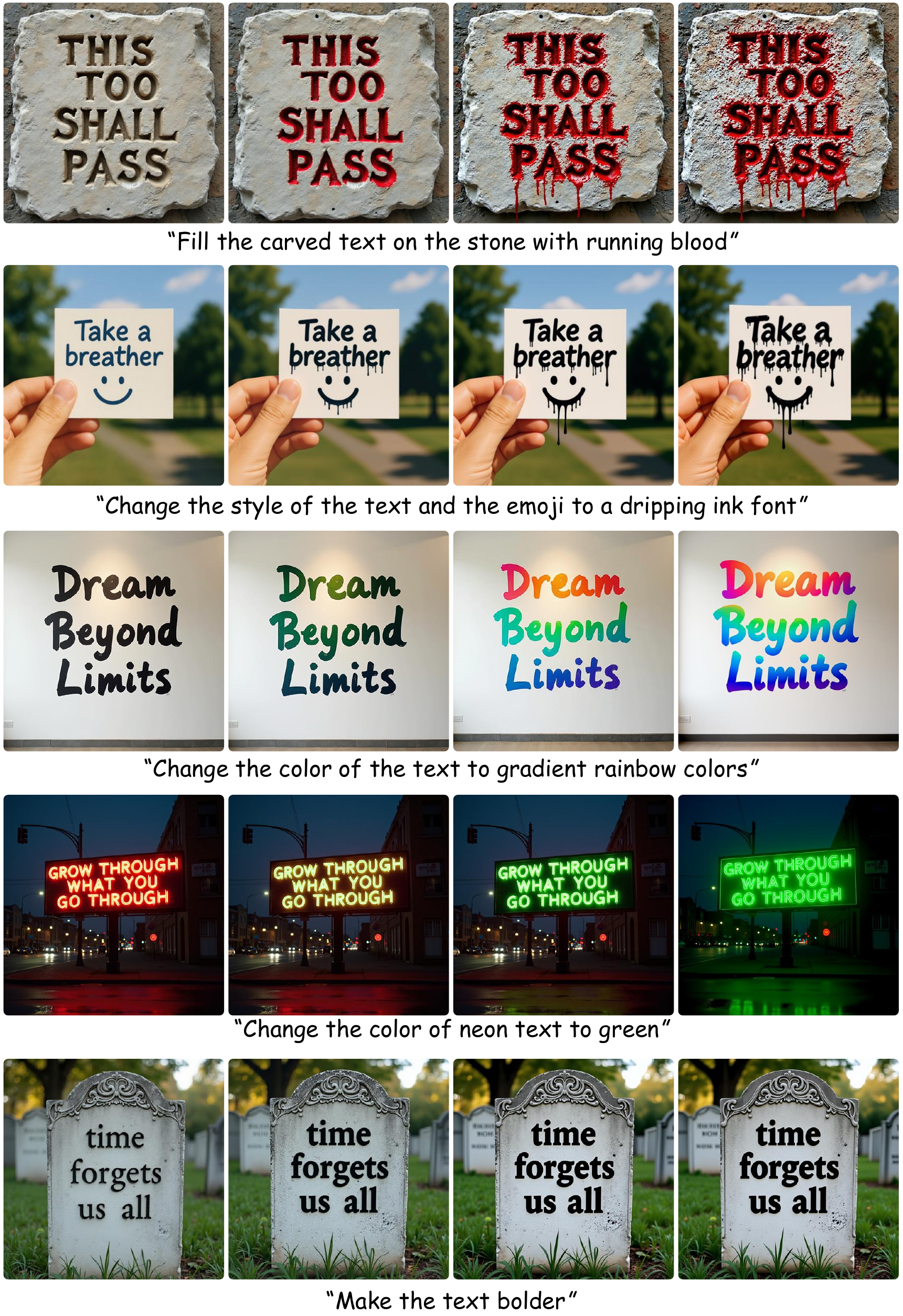}
    \caption{Qualitative results of GSTLoRA on text editing.}
    \label{fig:gstlora_text_slider_appendix}
\end{figure}

Diffusion models belong to a class of generative models based on stochastic differential equations (SDE). The core idea is to gradually corrupt data by adding noise through a stochastic forward process until the original data distribution becomes a simple Gaussian distribution. This process can be described as: 
$$dx = f(x, t) dt + g(t)dW_t,$$ 
where $f(x, t)$ denotes the drift term, $g(t)$ represents the diffusion coefficient, and $dW_t$ is the Wiener process (an infinitesimal step of Brownian motion, representing a small random Gaussian perturbation). The model then learns the reverse process, which reconstructs the original data distribution from pure noise. Mathematically, this reverse-time SDE is written as:
$$dx = [f(x, t) - g^2(t)\nabla_x\log p_t(x)]dt + g(t)dW_t,$$ 
where $\nabla_x \log p_t(x)$ is the score function, representing the gradient of the log-density of the data distribution at time $t$. Intuitively, the score function tells the model in which direction to move each noisy sample to recover the data distribution. In practice, diffusion models are trained to approximate this score function using a neural network $s_\theta(x, t)$. Training minimizes the score matching loss, defined as:
$$\mathbb{E}_{t \sim U(0, T), x \sim p_t(x)}[\lambda(t)||\nabla_x\log p_t(x)-s_\theta(x,t)||^2],$$
where $\lambda(t)$ is a time-dependent weighting function. Once trained, the model can sample new data by simulating the learned reverse process starting from Gaussian noise.

Flow matching methods are closely related to diffusion models, designed for training Continuous Normalizing Flows.
The key idea is to learn a deterministic transformation that maps an initial noise distribution to the target data distribution by integrating an ordinary differential equation (ODE). The evolution of a sample $x$ over time is governed by a time-dependent vector field $v_\theta(x,t)$, defined as:
$$\frac{dx}{dt}=v_\theta(x,t),$$
where $v_\theta(x,t)$ is a neural network parameterizing the vector field to be learned.
Training involves aligning this learned field with a predefined target vector field $v_t(x)$, which describes how samples should flow from noise to data at each time step. This is achieved by minimizing the flow matching loss:
$$\mathbb{E}_{t \sim U(0, T), x \sim p_t(x)}[|v_\theta(x,t)-v_t(x)|^2],$$
where $p_t(x)$ represents intermediate distributions along the transformation path from the initial to the final data distribution.
Unlike diffusion models, which rely on stochastic SDE trajectories involving random noise, flow matching employs deterministic ODE trajectories. This eliminates the stochasticity in sampling and generally leads to faster and more efficient training and inference. As a result, flow matching can be viewed as a computationally efficient deterministic counterpart to diffusion models.

\section{SliderEdit: Continuous Image Editing}
\label{sec:app:method}

\subsection{Simplified Partial Prompt Suppression Loss}
\label{sec:app:spps}

While the main Partial Prompt Suppression (PPS) objective requires selectivel suppressing an individual instruction $\mathcal{P}_i$ within a composite prompt $\mathcal{P}$, the Simplified PPS (SPPS) variant adopts a more streamlined approach that reduces this complexity a bit while maintaining strong generalization.

In SPPS, each training sample is treated as a single-instruction editing instance, i.e., $\mathcal{P} = \{\mathcal{P}_1\}$. The model learns to suppress the visual influence of this sole instruction by minimizing the difference between the denoising prediction of the adapted model when conditioned on $\mathcal{P}_1$ and that of the frozen base model when the prompt is removed entirely (or replaced with a prompt that acts as a null instruction, e.g., "keep the image the same"). Formally, the loss follows the same structure as $\mathcal{L}_\texttt{PPS}$:
\[
\mathcal{L}_\texttt{SPPS} = \|\epsilon_{M_\theta(\mathcal{P}_1)}(Z, X_\text{orig}, \mathcal{P}_1) -\epsilon(Z, X_\text{orig}, \varnothing\}) \|
\]
This encourages the adapter to learn how to neutralize the edit induced by $\mathcal{P}_1$, thereby isolating its corresponding representation within the model. Figure~\ref{fig:method_simple} visualizes the SPPS training pipeline.

Despite its simplicity, SPPS offers several practical advantages. It removes the need to parse multi-instruction prompts or identify token-level boundaries between sub-instructions, allowing efficient training on general instruction-based editing datasets, including those containing only single-instruction pairs. Moreover, the adapters trained with SPPS exhibit strong robustness and compositional generalization, performing effectively even when applied to multi-instruction edits at inference time. However, PPS provides finer-grained supervision, leading to more disentangled and well-localized adaptations across different instruction dimensions, which results in better control when handling complex multi-instruction edits.

\begin{figure*}[t]
    \centering
    \includegraphics[width=\linewidth]{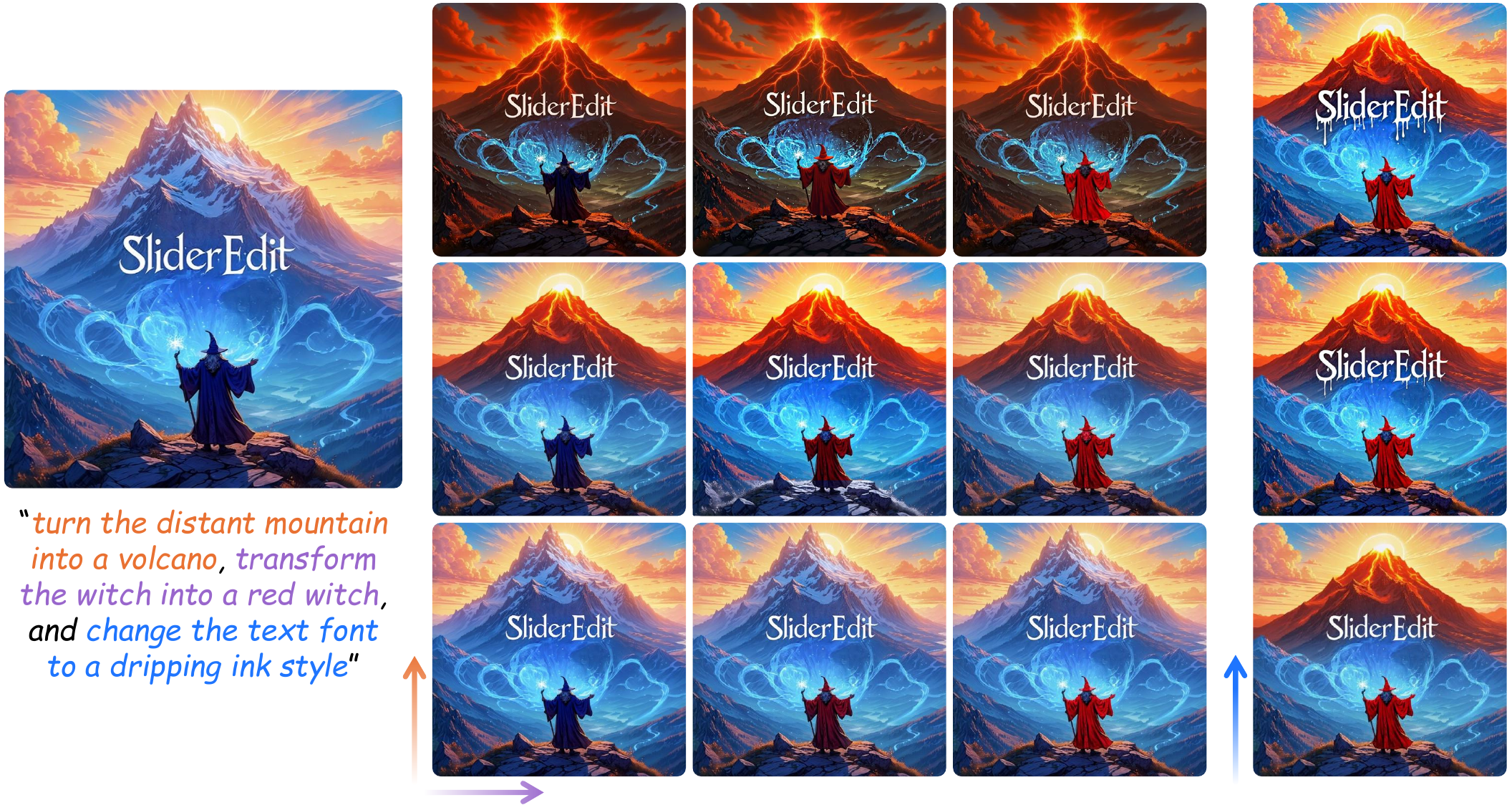}
    \caption{\textbf{Qualitative results of STLoRA on a 3-instruction edit.} The model demonstrates smooth and continuous control over the strength of each instruction in a disentangled manner.}
    \label{fig:app:stlora_3d_qualitative}
\end{figure*}

\begin{figure}[t]
    \centering
    \includegraphics[width=\linewidth]{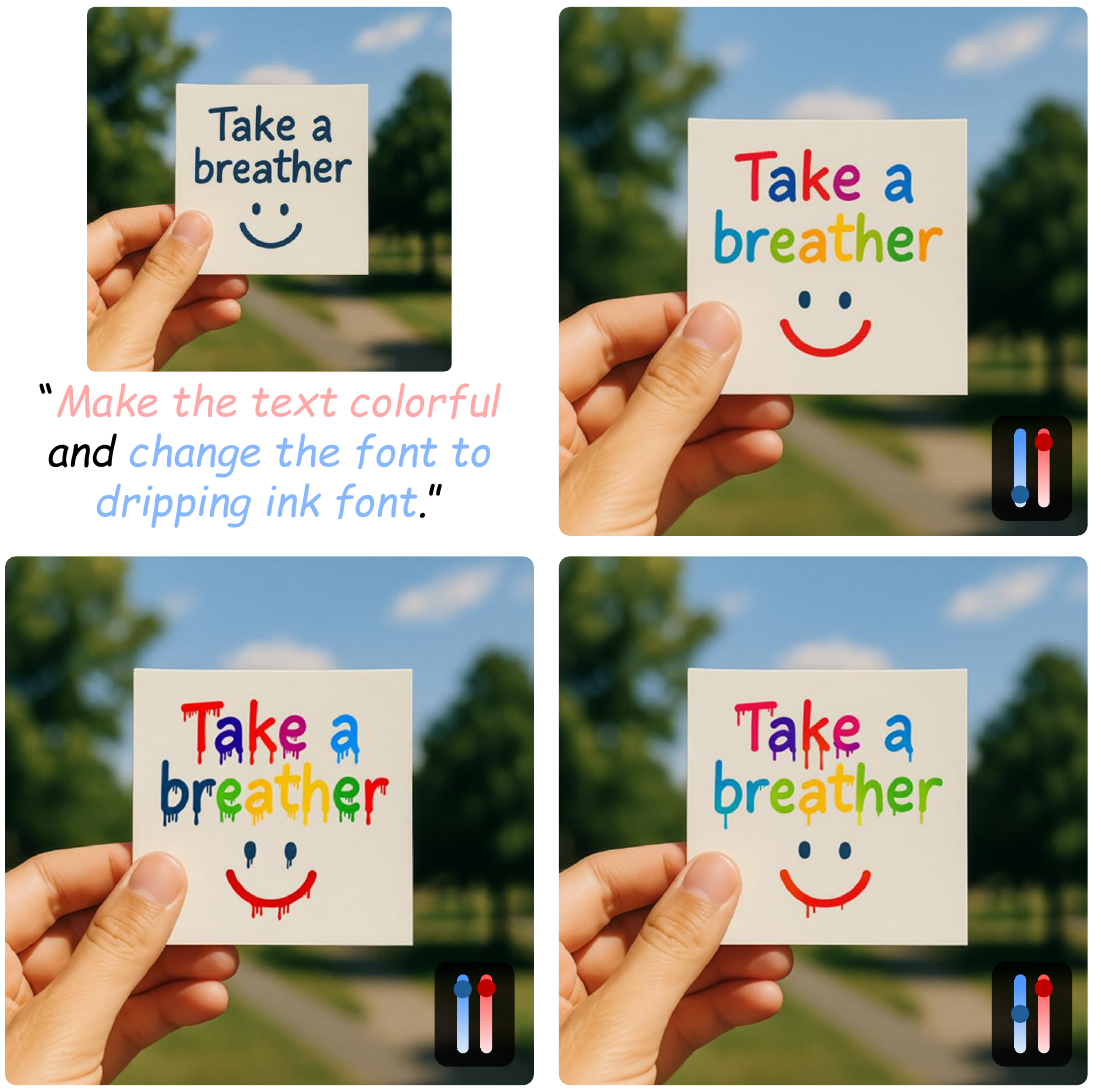}
    \caption{Qualitative results of STLoRA on a 2-instruction edit for text editing.}
    \label{fig:2d_stlora_text_slider}
\end{figure}

\section{Experiments}
\label{sec:app:experiments}

\subsection{Implementation Details}
\label{sec:app:implementation_details}

We use FLUX-Kontext and Qwen-Image-Edit\footnote{We adopt Qwen-Image-Edit-2509, an updated version with improved performance and stronger identity preservation.} as our base models. All models are trained with the $\ell_{\text{SPPS}}$ loss, chosen for its simplicity, efficiency, and strong generalization. We observe that $\mathcal{L}_{\text{PPS}}$ provides more robust and disentangled control for multi-instruction setups when used with STLoRA (see Appendix~\ref{sec:app:experiments_pps_vs_spps}). Training is performed on a small subset (1k–8k samples) of the GPT-Image-Edit-1.5M dataset~\cite{wang2025gpt}. For STLoRA, we train both base models for 1,000 iterations with a batch size of 8, observing early convergence around iterations 400–500 but continuing to 1,000 for consistency. For GSTLoRA, we train FLUX-Kontext for 300 iterations with a batch size of 4. We employ the AdamW optimizer with a learning rate of $1\times10^{-4}$, no warm-up, and train across all diffusion timesteps. All experiments are conducted on a single NVIDIA H100-SXM GPU using mixed-precision (\texttt{bfloat16}) training with gradient checkpointing for memory efficiency. The LoRA modules have a rank of 16 and zero dropout, and are applied to the $Q$, $K$, $V$, and output projections of the attention layers, as well as to the two additional linear projections in each transformer block. These settings provide a stable and memory-efficient training setup, enabling rapid convergence across all models. Overall, our training is computationally highly lightweight and data-efficient. Furthermore, consistent with prior observations in~\cite{esser2024scaling, zarei2025localizing}, we found that training adapters on only a subset of transformer blocks can achieve performance comparable to training all blocks. Also, following insights from~\cite{zarei2024improving, zarei2024understanding}, applying adapters at every denoising timestep may not be necessary for effective editing. We leave a comprehensive investigation of these efficiency-oriented design choices for future work.

\subsection{Quantitative  Results}
\label{sec:app:quantitative}

\subsubsection{Metrics}
\label{sec:app:metrics}

\textbf{Continuity.} 
Given a sequence of similarity scores $\{s_1, \ldots, s_\delta\}$ corresponding to increasing $\alpha$ values, we expect these scores to change smoothly and approximately uniformly between $\min(s_i)$ and $\max(s_i)$.  
To quantify this, we compute a chi-squared statistic,
\[
\chi^2 = \sum_{i=1}^{\delta} \frac{(O_i - E)^2}{E},
\]
where $O_i$ denotes the observed count in each bin (number of similarity scores $s_j$ falling within the $i$-th bin), and $E$ denotes the expected count per bin under a uniform distribution ($E=1$).
We report $({\chi^2_{\text{agg}}}/{\text{dof}})^{-1}$ ($\text{dof}:$degrees of freedom) as our continuity metric—larger values indicate higher continuity and smoother edit trajectories. For 2D and 3D edit spaces, we apply an analogous chi-squared test to evaluate the uniformity of the sample distribution across the corresponding grids.

\textbf{Disentanglement.}
To evaluate disentanglement, we measure how well the model isolates the intended edit without affecting unrelated aspects, such as identity or background.  
First, we assess \emph{identity preservation} using cosine distance in the identity embedding space obtained from \emph{ArcFace}~\cite{deng2019arcface}, where lower distances indicate stronger identity consistency.  
To capture more general visual changes, we compute feature distances between edited images $\mathcal{I}_i$ with the origin image using multiple perceptual metrics:  
\emph{LPIPS}~\cite{zhang2018unreasonable} (using both AlexNet \cite{NIPS2012_c399862d} and VGG \cite{simonyan2014very} backbones) and \emph{DINOv2}~\cite{caron2021emerging, oquab2023dinov2}.  
While LPIPS focuses on low-level perceptual similarity, DINO captures higher-level semantic consistency, allowing us to evaluate both appearance-level and structural disentanglement.

\begin{figure*}[t]
    \centering
    \includegraphics[width=\linewidth]{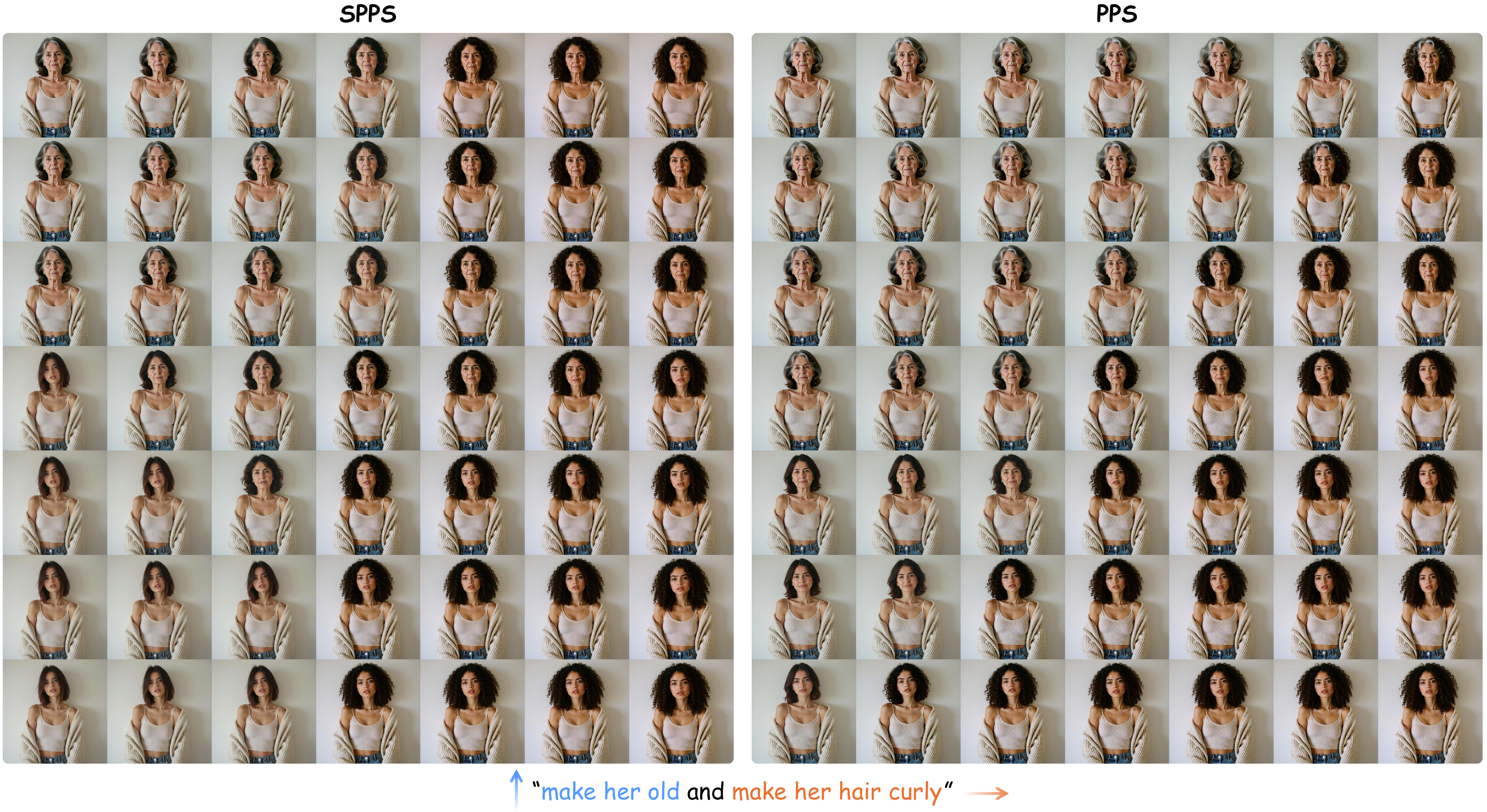}
    \vspace{-0.6cm}
    \caption{\textbf{Qualitative Comparison between PPS and SPPS.}
PPS produces a more disentangled and smoother interpolation space in multi-instruction editing scenarios, offering finer control over individual instruction directions compared to SPPS.}
    \label{fig:app:pps_vs_spps}
\end{figure*}

\subsubsection{Baselines}
\label{sec:app:baselines}
We consider different baselines depending on the number of edit instructions $\gamma$ used in the prompt. 

For the case of a single-instruction setting ($\gamma = 1$), we compare \emph{GSTLoRA} (Ours) and \emph{STLoRA} (Ours) with \emph{Explicit CFG} and \emph{Implicit CFG}, all implemented on top of the \emph{FLUX-Kontext} model, as well as Concept-Slider \cite{g2023concept} and Continuous Attribute Control \cite{baumann2024continuous}.  
Implicit CFG refers to the classifier-free guidance (CFG) mechanism applied in an implicit manner.  
FLUX-Kontext is a \emph{guidance-distilled} model, meaning that at inference time it does not explicitly perform CFG as:
\[
\epsilon_{\text{CFG}} = \epsilon_{\text{uncond}} + s \big( \epsilon_{\text{cond}} - \epsilon_{\text{uncond}} \big),
\]
where $\epsilon_{\text{cond}}$ and $\epsilon_{\text{uncond}}$ denote the conditional and unconditional predictions, respectively, and $s$ is the guidance scale. 
Instead, the model internally learns to approximate the effect of a given $s$, allowing us to vary this parameter to implicitly control guidance strength. However, as observed in our experiments, this implicit scaling provides only limited control over the edit intensity.

To enable explicit guidance, we first set the model's internal (implicit) guidance scale to $s = 1$, effectively recovering the base (unguided) model.  We then apply explicit CFG during inference using $\epsilon' = \epsilon_{\text{uncond}} + w \big( \epsilon_{\text{cond}} - \epsilon_{\text{uncond}} \big),$
where $w$ is the external CFG scale. This requires two forward passes through the model—one with the conditioning prompt and one without".

Concept-Slider and Continuous Attribute Control enable fine-grained attribute manipulation in text-to-image models. While they can be adapted to image editing via inversion methods \cite{mokady2023null, garibi2024renoise}, their performance in this setting is comparatively limited.

For cases involving multiple edit instructions ($\gamma > 1$), Explicit CFG, Implicit CFG, and GSTLoRA cannot independently control individual edit directions.  
This limitation highlights the advantage of \emph{STLoRA}, which enables disentangled, per-instruction control in multi-instruction editing scenarios. As Concept-Slider and Continuous Attribute Control show limited effectiveness even for single-instruction edits ($\gamma = 1$), we omit them from this setting.  
We evaluate STLoRA using both \emph{FLUX-Kontext} and \emph{Qwen-Image-Edit} models.

\begin{figure*}[t]
    \centering
    \includegraphics[width=\linewidth]{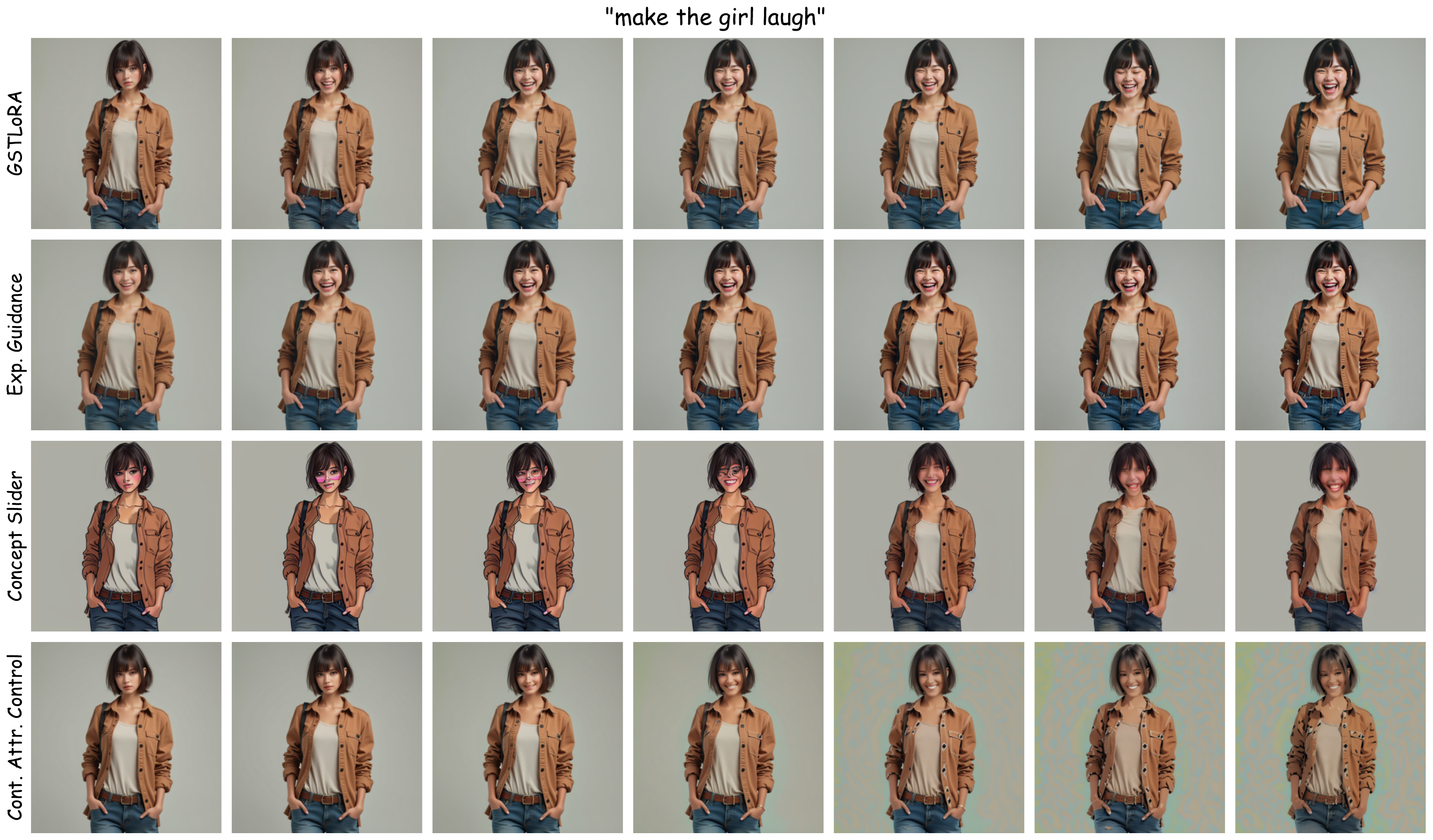}
    \caption{\textbf{Qualitative Comparison with Baselines.}
While SliderEdit (GSTLoRA variant here) and Explicit Guidance produce high-quality edits, Concept-Slider and Continuous Attribute Control perform poorly on real image editing, as they are primarily designed for text-to-image generation and rely on indirect inversion-based adaptation.}
    \label{fig:app:qualitative_comparison_with_baselines}
\end{figure*}

\subsection{Qualitative Results}
\label{sec:app:qualitative}

We provide additional qualitative results to further illustrate the capabilities of \emph{SliderEdit} and its variants across a diverse range of editing tasks.

\begin{figure*}[t]
    \centering
    \includegraphics[width=0.9\linewidth]{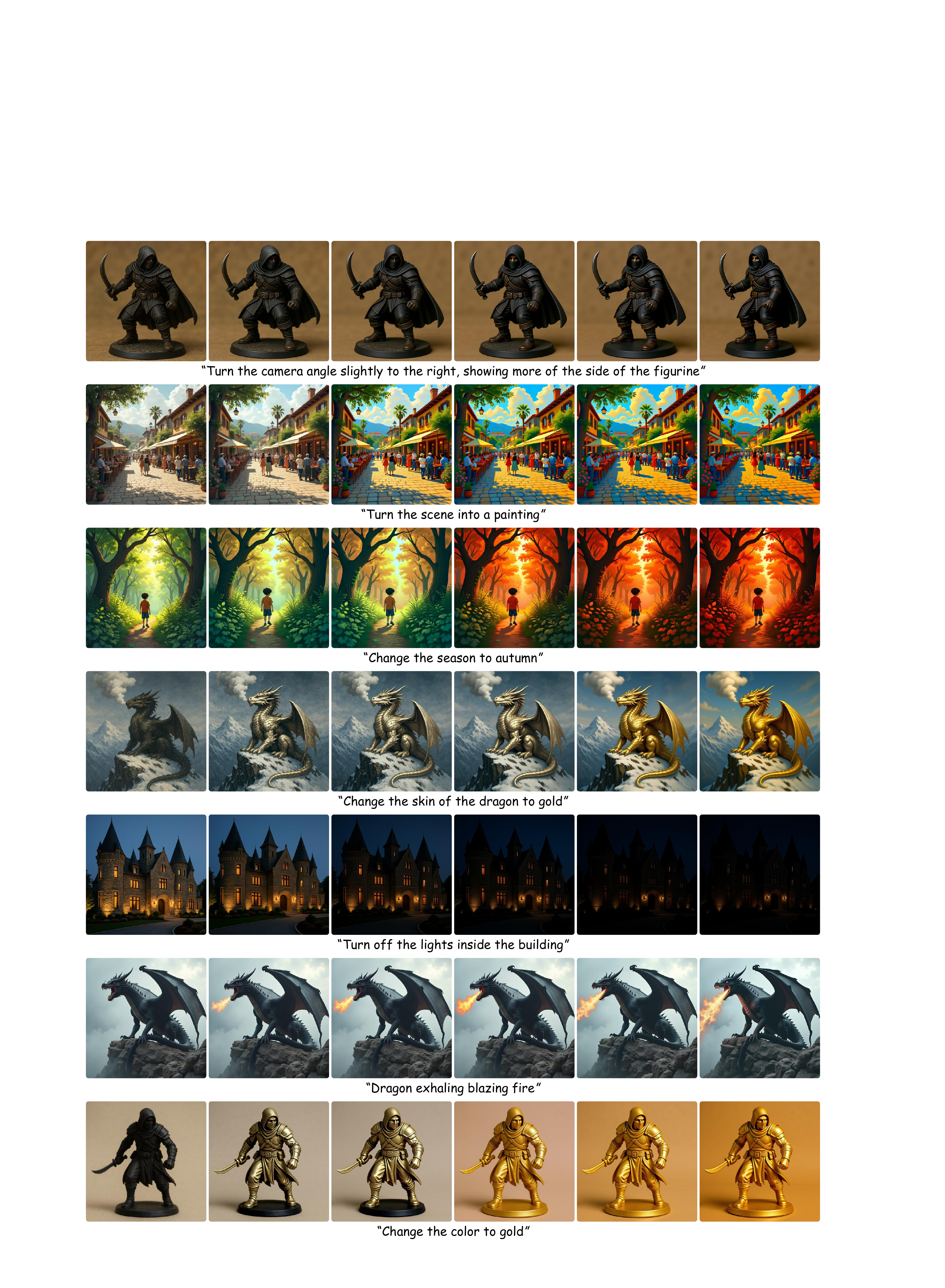}
    \caption{\textbf{Qualitative Samples of GSTLoRA.} The model emonstrates smooth, continuous control over the strength of both local and global edits.}
    \label{fig:app:gstlora_qualitative}
\end{figure*}

Figure~\ref{fig:app:gstlora_qualitative} showcases diverse examples generated using \emph{GSTLoRA}, demonstrating smooth and continuous control over both local and global edits. The model effectively interpolates between different edit strengths, producing coherent intermediate images without abrupt transitions.

\begin{figure*}[t]
    \centering
    \includegraphics[width=0.93\linewidth]{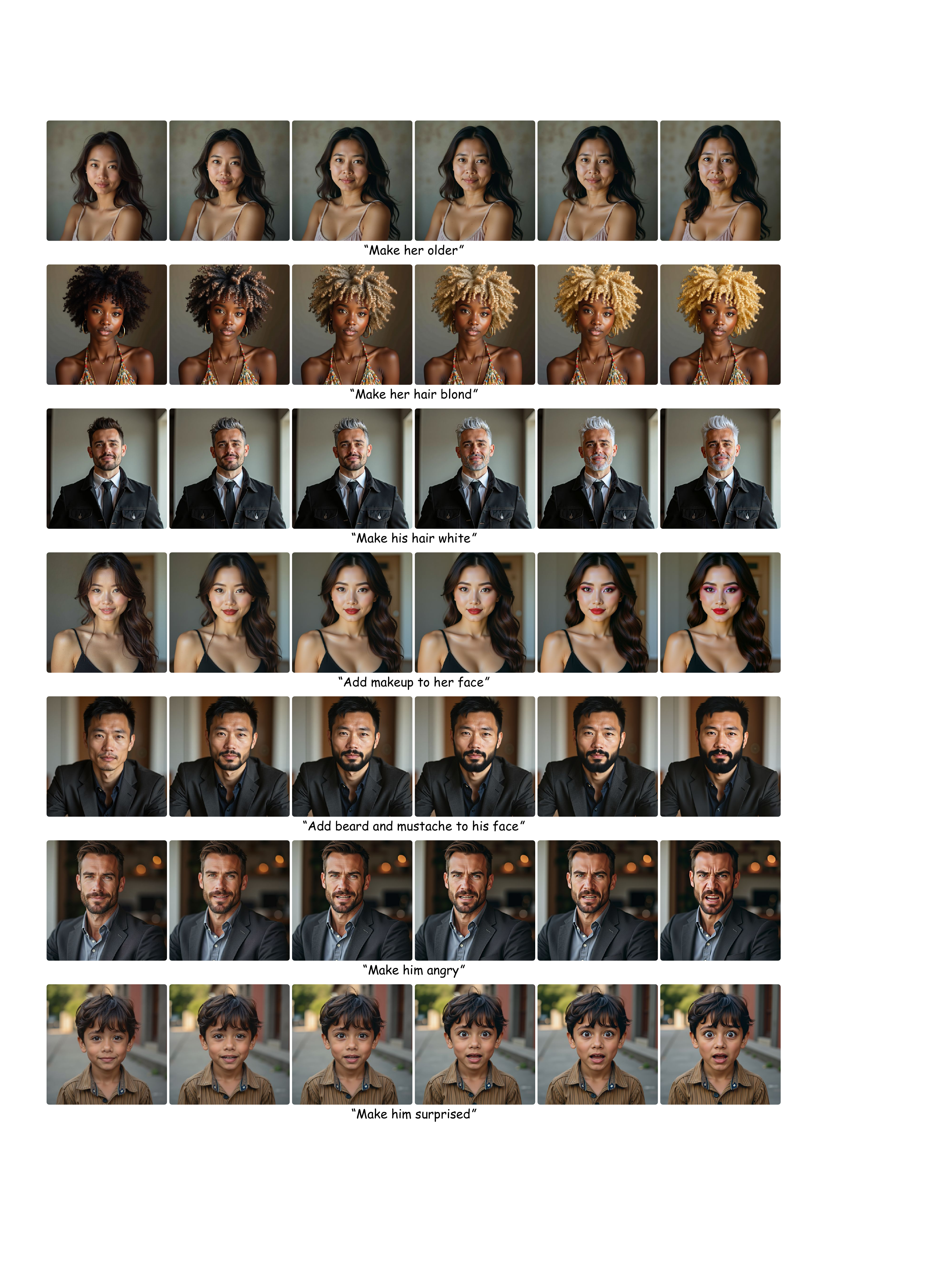}
    \caption{Qualitative results of GSTLoRA on face editing}
    \label{fig:app:gstlora_qualitative_face_1}
\end{figure*}

\begin{figure*}[t]
    \centering
    \includegraphics[width=0.93\linewidth]{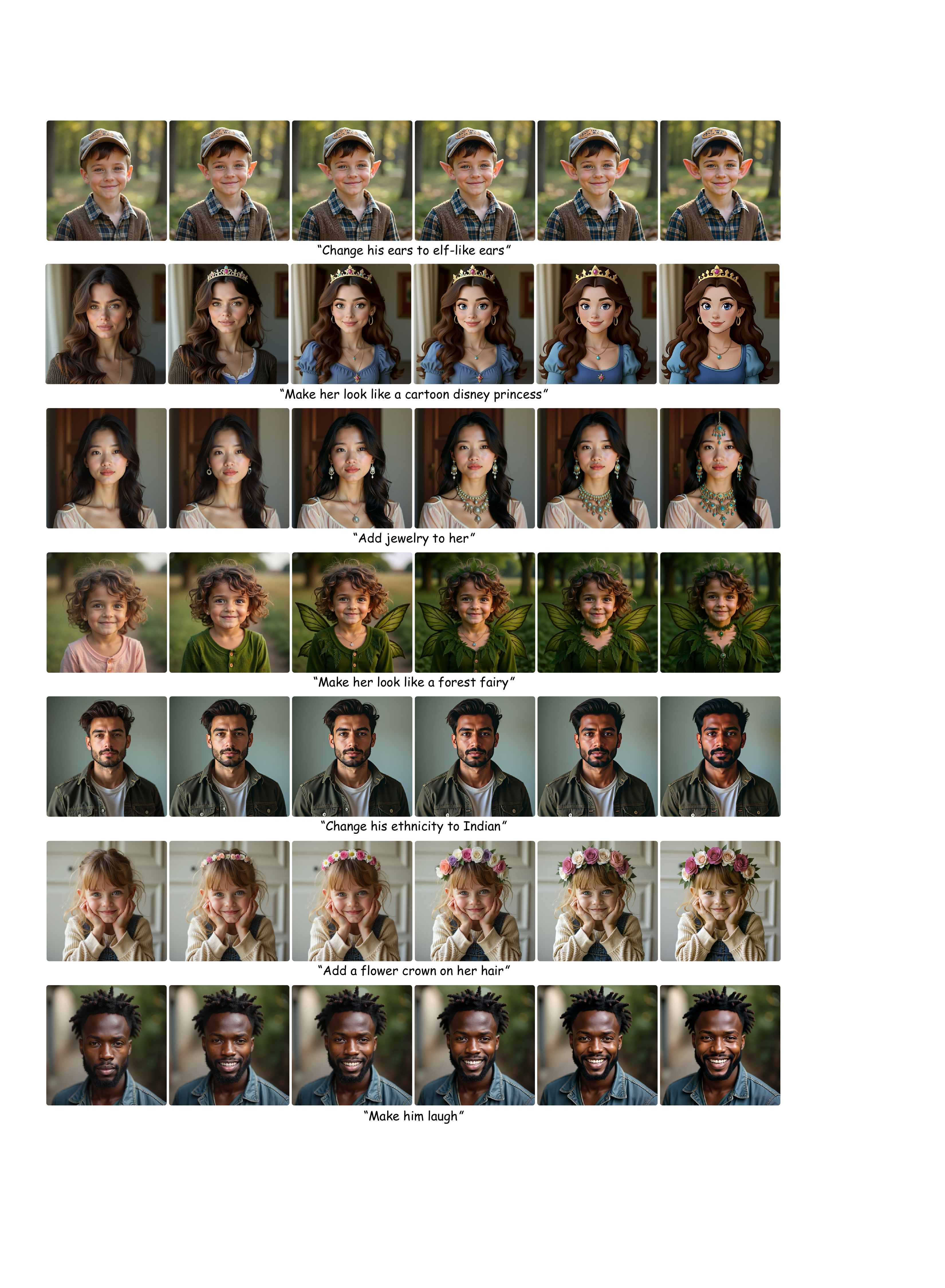}
    \caption{Qualitative results of GSTLoRA on face editing}
    \label{fig:app:gstlora_qualitative_face_2}
\end{figure*}

To further evaluate its capability in fine-grained manipulation, Figures~\ref{fig:app:gstlora_qualitative_face_1} and~\ref{fig:app:gstlora_qualitative_face_2} present qualitative results on face-editing tasks. The model can accurately and continuously adjust facial attributes such as hair length, curliness, makeup, skin tone, hair color, and age, as well as facial expressions including smiling, anger, and surprise. In addition, Figure~\ref{fig:gstlora_text_slider_appendix} demonstrates GSTLoRA’s versatility in \emph{text editing}. The model enables continuous adjustment of textual attributes such as font color, style, and weight.

\begin{figure*}[t]
    \centering
    \includegraphics[width=\linewidth]{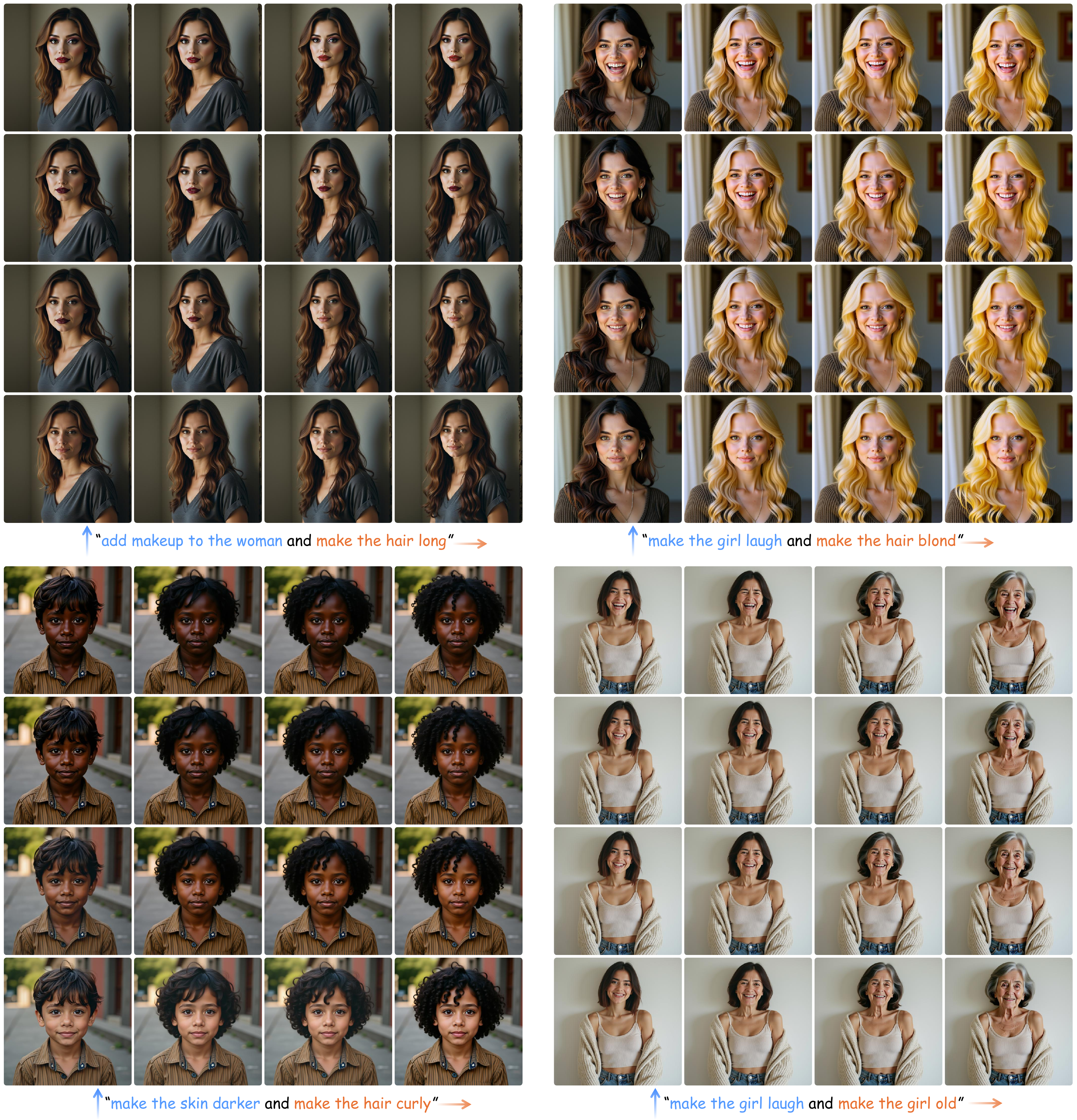}
    \caption{\textbf{Qualitative results of STLoRA on 2-instruction edits.} The model demonstrates smooth, continuous control over the strength of both directions.}
    \label{fig:app:stlora_2d_qualitative}
\end{figure*}

Figures~\ref{fig:app:stlora_2d_qualitative} and~\ref{fig:app:stlora_3d_qualitative} illustrate qualitative results of \emph{STLoRA} on multi-instruction editing tasks. In the 2-instruction setting, the model produces a smooth and interpretable 2D interpolation space, where each axis corresponds to a distinct instruction direction. Extending this to 3-instruction scenarios (Figure~\ref{fig:app:stlora_3d_qualitative}), STLoRA maintains disentangled control, allowing continuous modulation of each instruction independently.
We further demonstrate STLoRA’s capability on \emph{text editing} tasks in Figure~\ref{fig:2d_stlora_text_slider}, where the model learns disentangled control over multiple text attributes (e.g., font style and color).

Overall, these results highlight the flexibility and generality of the proposed framework across domains, showing that both GSTLoRA and STLoRA enable smooth, continuous, and disentangled control over diverse editing operations.

\subsubsection{PPS vs SPPS}
\label{sec:app:experiments_pps_vs_spps}
We compare the Partial Prompt Suppression (PPS) and Simplified PPS (SPPS) objectives to assess their effect on disentanglement and control quality. As illustrated in Figure~\ref{fig:app:pps_vs_spps}, both objectives enable smooth and continuous interpolation along edit directions. However, PPS produces a more disentangled latent space, allowing finer and more independent control over each instruction, while SPPS serves as a simpler yet effective alternative that achieves comparable results in most cases.

Some degree of attribute entanglement persists across all models, including the underlying base model. For instance, even when using the base instruction-based editing model, modifying a person’s skin tone can unintentionally affect correlated features such as hair color or lighting. This behavior arises from inherent attribute coupling in the generative model itself, rather than from limitations introduced by our sliders.

\subsubsection{Comparison with other baselines}
As shown quantitatively in Table~\ref{tab:1d_comparison} and discussed in Section~\ref{sec:experiments}, Concept-Slider and Continuous Attribute Control perform poorly on real image editing tasks due to their indirect adaptation from text-to-image generation. Here, we provide qualitative examples in Figure~\ref{fig:app:qualitative_comparison_with_baselines} for visual comparison. While SliderEdit (GSTLoRA variant in this case ) and Explicit Guidance produce smooth, coherent, and faithful edits aligned with the input instructions, Concept-Slider and Continuous Attribute Control often fail to maintain image fidelity or accurately follow the target modification. These qualitative results further confirm the quantitative findings, demonstrating that SliderEdit enables both fine-grained control and high-quality real image editing.

\end{document}